%% file: main.tex
\def\wacvPaperID{666} %
\def\confName{WACV}
\def\confYear{2026}

\documentclass[10pt,twocolumn,letterpaper]{article}

\usepackage[pagenumbers]{wacv} %

\DeclareMathOperator{\lcm}{lcm}

\usepackage{scalerel}
\usepackage{bigfoot}
\usepackage{multirow}
\usepackage{booktabs}
\usepackage{makecell}
\usepackage{amssymb}  %
\usepackage{pifont}   %
\usepackage{dirtytalk}
\usepackage{caption}
\usepackage[T1]{fontenc}
\usepackage{enumitem}
\usepackage[switch]{lineno}
\usepackage{float}
\usepackage[table,dvipsnames]{xcolor}
\usepackage{adjustbox}
\usepackage{changepage}
\usepackage{fontawesome5}
\usepackage{MnSymbol}
\usepackage{stmaryrd}
\usepackage{tikz,graphicx}
\usepackage{stfloats}  %

\newcommand{\greyrule}{\arrayrulecolor{black!30}\midrule\arrayrulecolor{black}}

\newcommand{\cmark}{\textcolor{Green}{\ding{51}\xspace}}%
\newcommand{\xmark}{\textcolor{Red}{\ding{55}}\xspace}%

\newcommand{\yes}{\textcolor{black!50}{\faCheck}}%
\newcommand{\no}{\textcolor{black!50}{\faTimes}}%
\newcommand{\mono}{\textcolor{black!50}{\faCalendarDay}}%
\newcommand{\multitemp}{\textcolor{black!50}{\faCalendarWeek}}%
\newcommand{\phantomeye}{\phantom{\faEye}}%
\newcommand{\eyeslash}{\textcolor{black!50}{\faEyeSlash}}

\newcommand{\greenmultitemp}{\textcolor{ForestGreen!75}{\faCalendarWeek}}%
\newcommand{\greeneye}{\textcolor{ForestGreen!75}{\faEye}}

\newcommand{\early}{\textcolor{black!50}{\faHourglassStart~}}%
\newcommand{\inter}{\textcolor{black!50}{\faHourglassHalf~}}%
\newcommand{\tbased}{\textcolor{black!50}{\faObjectUngroup~}}%
\newcommand{\jointt}{\textcolor{black!50}{\faObjectGroup~\hspace{1px}}}%
\newcommand{\share}{\textcolor{black!50}{\faRetweet~}}%

\usepackage{endnotes}
\let\footnote=\endnote
\makeatletter
\renewcommand{\paragraph}{%
  \@startsection{paragraph}{4}%
  {\z@}{1.ex \@plus 1ex \@minus .2ex}{-1em}%
  {\normalfont\normalsize\bfseries}%
}
\makeatother

\input{preamble}
\definecolor{wacvblue}{rgb}{0.21,0.49,0.74}
\usepackage[pagebackref,breaklinks,colorlinks,allcolors=wacvblue]{hyperref}

\usepackage[capitalize]{cleveref}
\crefname{section}{Sec.}{Secs.}
\Crefname{section}{Section}{Sections}
\Crefname{table}{Table}{Tables}
\crefname{table}{Tab.}{Tabs.}

\def\maketitlesupplementarysingle{
\clearpage %
  \onecolumn  %
  \thispagestyle{plain} %
  \begin{center}
    \Large
    \textbf{\thetitle}\\
        \vspace{0.5em}Supplementary Material \\
        \vspace{1.0em}
  \end{center}
}
\title{MAESTRO: Masked AutoEncoders for Multimodal, Multitemporal, and Multispectral Earth Observation Data}

\author{
    Antoine Labatie $^{1}$ \quad
    Michael Vaccaro $^{1}$ \quad
    Nina Lardiere $^{1}$ \quad
    Anatol Garioud $^{1}$ \quad
    Nicolas Gonthier $^{1,2}$ \\[4pt]
    $^{1}$ Institut national de l’information géographique et forestière (IGN), France\\[4pt]
    $^{2}$ Univ Gustave Eiffel, ENSG, IGN, LASTIG, France\\[4pt]
    {\small \texttt{\{firstname.lastname\}@ign.fr}}
}

\begin{document}

\twocolumn[{
\maketitle
\vspace{-10mm}
\begin{center}
    \captionsetup{type=figure}
    \begin{tikzpicture}
  \node[opacity=1] {\includegraphics[width=\linewidth]{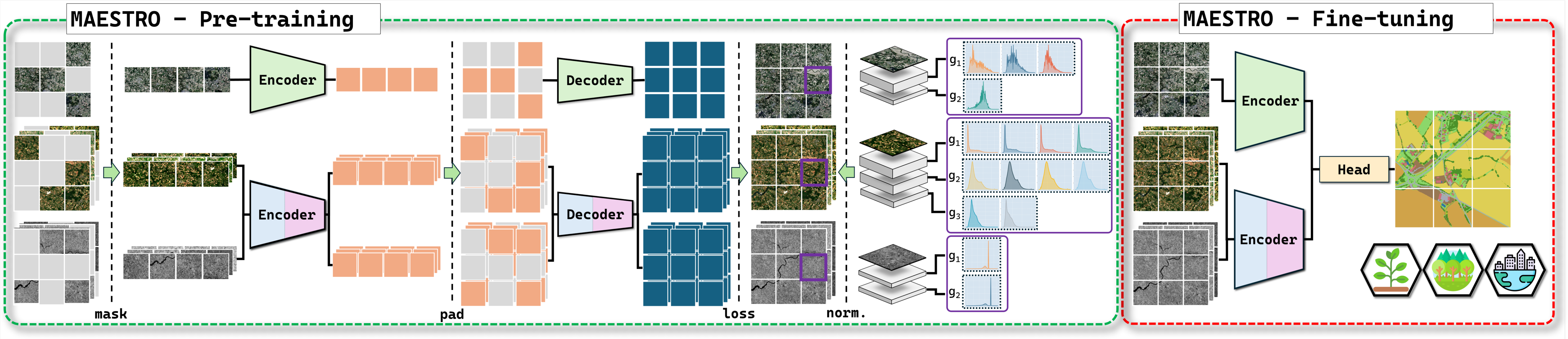}};
\end{tikzpicture}
    \captionof{figure}{\textbf{Overview of MAESTRO}. MAESTRO extends the Masked Autoencoder to orchestrate the complex interplay of multimodal, multitemporal, and multispectral Earth Observation data. It employs token-based early fusion across time steps and similar modalities, and token-based late fusion across dissimilar modalities. It uses joint-token fusion for multispectrality, but still relies on a novel normalization of reconstruction targets—namely, patch-group-wise within groups of highly correlated bands—to inject a useful spectral prior during pre-training. Best viewed in color.}
    \label{fig:overview}
\end{center}
}]

\input{0_abstract}    
\input{1_intro}
\input{2_related_work}
\input{3_approach}

\input{4_experiments}

\input{5_conclusion}

\newpage
\clearpage

\section*{Acknowledgement} This work was granted access to the HPC/AI resources provided by GENCI-IDRIS (allocations A0181013803, A0161013803, and AD010114597R1). We thank SIMV/SDM colleagues at IGN, Guillaume Astruc, Loïc Landrieu, Alexandre Tuel, and Thomas Kerdreux for useful discussions.

{
    \small
    \bibliographystyle{ieeenat_fullname}
    \bibliography{main}
}
\input{6_experimental_details}
\input{7_detailed_results}

\input{8_add_results}

\input{9_inference}

\input{10_flops}

\end{document}

%% file: 0_abstract.tex
\begin{abstract}
Self-supervised learning holds great promise for remote sensing, but standard self-supervised methods must be adapted to the unique characteristics of Earth observation data. We take a step in this direction by conducting a comprehensive benchmark of fusion strategies and normalization schemes of reconstruction targets for multimodal, multitemporal, and multispectral Earth observation data. Based on our findings, we introduce MAESTRO, a novel adaptation of the Masked Autoencoder with optimized fusion mechanisms and a normalization scheme that incorporates a spectral prior as a self-supervisory signal. Evaluated on four Earth observation datasets in both intra- and cross-dataset settings, MAESTRO achieves state-of-the-art performance on tasks that strongly rely on multitemporal dynamics, while also remaining competitive on others.
Code to reproduce all our experiments is available at \url{https://github.com/ignf/maestro}.
\end{abstract}

%% file: 1_intro.tex
\vspace{-4mm}
\section{Introduction}
\label{sec:intro}
Self-supervised learning (SSL) has been central to recent breakthroughs in natural language processing \cite{radford2018,devlin2019} and computer vision \cite{he2021,xie2021}. It is an effective pre-training strategy, improving versatility and data efficiency when fine-tuning models on diverse downstream tasks—especially in the case of foundation models pre-trained on large amounts of unlabeled data.

The SSL paradigm is promising for Earth Observation (EO) applications, where unlabeled data is abundant, but labeled data remains scarce and costly. However, to fully leverage the potential of SSL, it may be necessary to adapt off-the-shelf approaches to the unique characteristics of EO data. A key characteristic of EO data lies in its heterogeneity across resolutions, scales, modalities, and spectral bands \cite{rolf2024}. This heterogeneity is frequently reflected in multimodal, multitemporal, and multispectral EO datasets, where the integration of various sensors effectively addresses the spatial-temporal-spectral \say{resolution dilemma} \cite{al2013}.

Our goal in this work is to efficiently adapt the canonical SSL framework of masked autoencoding to the specific characteristics of EO data. Concretely, we aim to design SSL methods that (i) effectively address the main axes of heterogeneity in EO data—multimodality, multitemporality, and multispectrality—and (ii) natively support very high-resolution (VHR) imagery.

To this end, we build on the Masked Autoencoder (MAE) \cite{he2021}, which has demonstrated strong performance and computational efficiency on natural images. Extending the MAE to multimodal, multitemporal, and multispectral EO data, however, poses several unique challenges.

First, integrating multiple modalities and temporal observations requires an effective fusion strategy. One approach, \emph{joint-token fusion}, projects all data dimensions into a shared token space. An alternative, \emph{token-based fusion}, first clusters data dimensions, embeds them separately, and fuses the resulting tokens. Token-based fusion appears more suited for handling multimodality/multitemporality as it (i) better captures the heterogeneity across modalities and time steps, (ii) allows the integration of modal and temporal encodings, and (iii) enables cross-modal and cross-temporal self-supervisory signals. Yet, it remains unclear whether fusion should be applied early or late in the model.

Second, handling multispectrality adds further complexity. Here, the choice between token-based and joint-token fusion is less obvious. While joint-token fusion offers computational efficiency, it limits the ability to incorporate cross-spectral self-supervisory signals. A key question is whether we can preserve the computational benefits of joint-token fusion while reintroducing an effective prior on the data's multispectral structure—potentially through a carefully designed normalization of reconstruction targets.

We investigate these questions in \cref{sec:ablation} and, based on our findings, we introduce our proposed approach, \textbf{MAESTRO}, in \cref{sec:maestro}. MAESTRO extends the MAE framework by applying token-based early fusion across time steps and similar modalities, but late fusion across dissimilar modalities. For handling multispectrality, MAESTRO employs joint-token early fusion combined with a novel normalization scheme for reconstruction targets.

Our main contributions are as follows:
\begin{itemize}
\item We conduct an extensive benchmark of fusion strategies and target normalization schemes for multimodal, multitemporal, and multispectral SSL in EO.
\item We introduce a novel patch-group-wise normalization method that normalizes reconstruction targets patch-wise within groups of highly correlated spectral bands. This approach injects a useful spectral prior into the self-supervised signal at negligible computational cost.
\item Building on the above, we propose MAESTRO, a tailored adaptation of the MAE framework that effectively orchestrates the use of multimodal, multitemporal, and multispectral EO data—both in terms of performance and computational cost.
\item We validate MAESTRO on four EO benchmarks, demonstrating that it achieves state-of-the-art results on tasks that strongly rely on multitemporal dynamics—both in intra- and cross-dataset transfers—while remaining highly competitive on others.
\end{itemize}

%% file: 2_related_work.tex
\section{Related Work}
\label{sec:related_work}

\paragraph{SSL Approaches and Models for EO.}
SSL has emerged as a promising approach for EO, due to the abundance of unlabeled remote sensing data. 
Early work on SSL for EO focused primarily on contrastive \cite{manas2021,ayush2021,jain2022,wang2022a,prexl2023,mall2023,wang2024c,wanyan2024,wang2024a,tian2024,danish2025,kerdreux2025}, often using convolutional neural networks \cite{manas2021,ayush2021,jain2022,prexl2023,mall2023,wanyan2024,wang2024a,tian2024}. Positive pairs were typically derived from different modalities, dates, or augmented views at the same location.
More recent works have increasingly shifted toward generative SSL using Transformers, well-suited to masked autoencoding. These include approaches inspired by MAE \cite{wang2022b,cong2022,lin2023,cha2023,jakubik2023,reed2023,tang2023,irvin2023,tseng2023,bountos2025,wang2024b,wang2024f,li2024c,zhang2024,hong2024,wang2024d,xiong2024,wang2024e,nedungadi2024,szwarcman2024,wang2025a,velazquez2025} or SimMIM \cite{sun2023,mendieta2023,han2024,diao2024,hu2024,spradlin2024}, or custom generative approaches \cite{smith2023,wang2023,dumeur2024,duc2025,wang2025b}. Some recent works also explore hybrid contrastive/generative approaches \cite{muhtar2023,dias2024} or latent-space generative SSL \cite{li2024a,tseng2025,waldmann2025,choudhury2025rejepa}.

\paragraph{Multimodal and Multitemporal SSL for EO.}
Recent studies in EO have begun to address multimodal \cite{wang2022a,jain2022,irvin2023,prexl2023,fuller2023,bountos2025,han2024,xiong2024,wang2024a,hu2024,nedungadi2024,prexl2024,jakubik2025,wang2025a,danish2025,waldmann2025,bi2025} or multitemporal SSL \cite{cong2022,jakubik2023,noman2024,dumeur2024,szwarcman2024}, but rarely both simultaneously. When multimodality is included, it is typically handled through parameter sharing or late fusion \cite{wang2022a,bastani2023,irvin2023,han2024,xiong2024,nedungadi2024,prexl2024,wang2025a}. To date, most published EO foundation models support multimodal or multitemporal inputs only via late fusion, which limits representation learning and downstream performance (see \cref{sec:ablation}).

Notable exceptions include Presto \cite{tseng2023}, Galileo \cite{tseng2025}, OmniSat \cite{astruc2024a}, AnySat \cite{astruc2024b}, SeaMo \cite{li2024b}, EarthMAE \cite{velazquez2025}, SkySense v1/v2 \cite{guo2024,zhang2025}, and SkySense++ \cite{wu2025}. Among these, Presto processes data pixel-wise without spatial context; Galileo applies token-based early fusion across all time steps and modalities; OmniSat and AnySat use joint-token early fusion via LTAE \cite{garnot2020} across time steps. 

In contrast, MAESTRO adopts token-based early fusion across time steps and within similar modalities only, thereby avoiding the inefficiencies of sharing encoders across dissimilar modalities. Despite the central role of fusion design in multimodal and multitemporal SSL (see \cref{sec:ablation}), to our knowledge only one prior work \cite{li2024b} has benchmarked different fusion strategies.

Beyond these limitations on the handling of multimodality/multitemporality, many existing approaches also lack explicit support for VHR data—a main goal of our work.

\paragraph{Multispectral SSL for EO.}
Various SSL studies in EO have incorporated SAR \cite{lin2023,hu2024,li2024a,kerdreux2025} or multispectral data \cite{manas2021,cong2022,smith2023,irvin2023,mendieta2023,jakubik2023,prexl2024,wanyan2024,spradlin2024,daroya2024,szwarcman2024,tian2024,dumeur2024} or both \cite{jain2022,wang2022a,prexl2023,fuller2023,bastani2023,tseng2023,nedungadi2024,han2024,wang2024d,wang2024a,wang2024c,xiong2024,astruc2024a,astruc2024b,bountos2025,wang2025a,tseng2025,waldmann2025,li2025a,li2025b,bi2025,danish2025,velazquez2025,jakubik2025}. However, only a few have explicitly encoded spectral priors into the model architecture or the self-supervisory signal. Some prior works have used token-based multispectral fusion by assigning separate tokens to each spectral band \cite{bountos2025,prexl2024} or to groups of spectral bands \cite{cong2022,irvin2023,tseng2023,tseng2025}. While these designs can help mitigate early information bottlenecks, enhance modeling capacity, and enable cross-spectral self-supervisory signals, they come at the cost of higher computational demands.

A handful of studies have explored the normalization of reconstruction targets for multispectral data, but their approaches have been restricted to patch-wise normalization without modeling the underlying spectral structure \cite{nedungadi2024,velazquez2025}. In contrast, we normalize reconstruction targets across groups of highly correlated spectral bands, effectively injecting a spectral prior into the self-supervisory signal. This approach enhances spectral representation learning without increasing the token count and naturally extends to SAR inputs via analogous band-grouping strategies.

\paragraph{Comparison Table.} SM \cref{tab:comparison_high_level} further positions our approach by detailing its main distinctions from prior work.

%% file: 3_approach.tex
\section{Approach}
\label{sec:approach}

\subsection{Architecture}
\label{sec:architecture}

In this section, we describe our MAE-derived approach specifically tailored to EO data. Our approach accommodates the heterogeneous nature of EO data, notably its multimodal, multitemporal, and multispectral characteristics.

We describe our approach in the context of a fixed dataset $\mathcal{D}$, composed of a set of modalities $m \in \mathcal{M}$. For each dataset tile, every modality $m$ is associated with an input tensor of shape $I_m \times I_m \times T_m \times C_m$, where $I_m$ denotes the spatial size, $T_m \geq 1$ the number of original time steps, and $C_m$ the number of channels. We assume that $C_m$ and $I_m$ remain constant across all tiles, while $T_m$ may vary. This variability reflects the nature of satellite time series, which may span a fixed duration but with location-dependent revisit intervals—such as with Sentinel-1 or Sentinel-2.

\paragraph{Multitemporal Discretization.} We start by enforcing a fixed tensor shape for the model inputs. For each modality $m$, we define a target number of discretized time steps $D_m$. For modality $m$, the original input tensor of shape $I_m \times I_m \times T_m \times C_m$ is reduced to a tensor of shape $I_m \times I_m \times D_m \times C_m$ through a two-step process: (i) temporal binning, where the sequence is reshaped into $D_m$ bins, each covering $T_m / D_m$ time steps; and (ii) time step selection within each bin.
During training, time step selection within each bin is random to introduce data augmentation. During validation and testing, the selection aims to maximize the representativeness of each bin. 
For more information, refer to SM \cref{sm:preprocessing}.

\paragraph{Patchification/Unpatchification.} Once the inputs have been reduced to a fixed shape, they are passed to modality-specific tokenizers. The tokenizers are shared across time steps within a given modality but \emph{distinct} across modalities. Tokenization is applied independently at each time step.

We adopt the standard Vision Transformer (ViT) patchification strategy \cite{dosovitskiy2020}, where for any modality $m$ and time step, inputs are first partitioned spatially into non-overlapping patches of size $P_m$. Each patch is then flattened into a vector of shape $P_m^2 C_m$ and projected into a patch embedding of dimension $C_e$ before the encoder. All spectral bands are jointly projected into the same tokens, implementing \emph{joint-token multispectral fusion}. During SSL pre-training, we apply transposed unpatchification to convert the decoded tokens of shape $C_d$ after the decoder back into reconstructed flattened patches of shape $P_m^2 C_m$.

For each modality $m$, the set of patch positions $p \in \mathcal{P}_m$ has cardinality $|\mathcal{P}_m|=(I_m / P_m)^2$, while the set of temporal bins $t \in \mathcal{T}_m$ has cardinality $|\mathcal{T}_m|=D_m$. Importantly, we do not assume that all modalities share the same number of spatial positions or temporal bins—that is, their spatial and temporal resolutions may differ. For example, modalities such as Sentinel-1 and Sentinel-2 may employ coarser spatial grids but finer temporal grids, whereas modalities such as aerial and SPOT 6–7 may use finer spatial grids with coarser temporal resolution. Maintaining these coarse or fine resolutions throughout tokenization helps prevent information bottlenecks in the tokenizers. %

Following tokenization, we add (i) two-dimensional spatial positional encodings based on ground sampling distance, as in ScaleMAE \cite{reed2023}, and (ii) sine-cosine temporal encodings, as in SatMAE \cite{cong2022}. We do not include explicit modality encodings, as modality-specific tokenizers and learnable modality-specific \verb|[mask]| tokens implicitly encode the source or target modality associated with each token. For more information, refer to SM \cref{sm:encodings}.

\paragraph{Multimodal and Multitemporal Token-Based Fusion.} At this stage, we obtain embedded inputs in the form of tensors of shape $I_m / P_m \times I_m / P_m \times D_m \times C_e$ for each modality $m$. As in the original MAE \cite{he2021}, these tensors are processed in two stages: a Transformer encoder processes only the visible tokens, while a Transformer decoder processes the encoded tokens concatenated with \verb|[mask]| tokens.

However, the original MAE workflow was designed for monomodal and monotemporal data, and there remains ambiguity on how it should be extended to support multimodal and multitemporal data. This raises key questions:
\begin{itemize}
\item Should information from different modalities and time steps be fused via \emph{early fusion} or \emph{late fusion}?
\item Should encoder and decoder parameters be shared across modalities or kept independent?
\end{itemize}

\begin{figure*}[ht]
    \centering
\includegraphics[width=\textwidth]{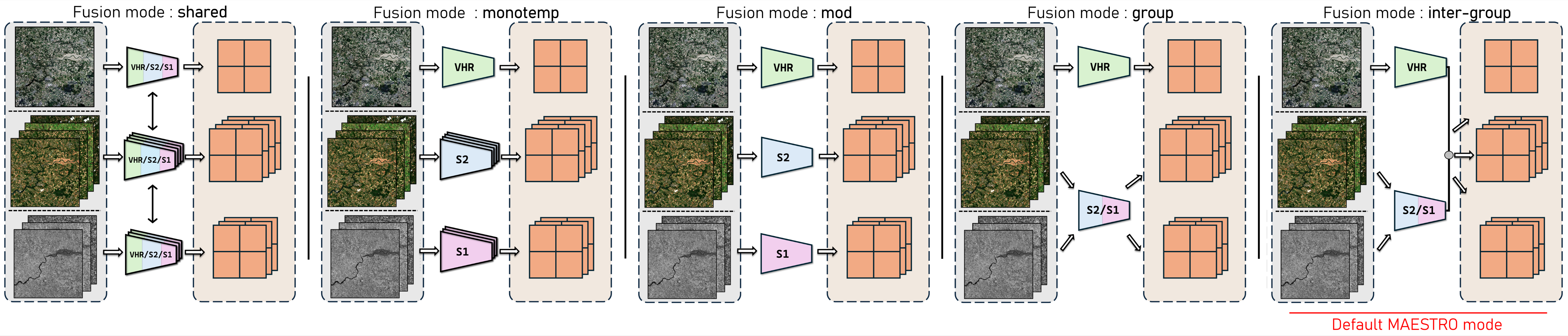}
    \caption{\textbf{Token-based fusion modes for handling multimodality and multitemporality.} Modes \emph{shared} and \emph{monotemp} involve late fusion across modalities and time steps, with parameters either shared across modalities (\emph{shared}) or kept independent for each modality (\emph{monotemp}). Mode \emph{group} involves late fusion across predefined groups of modalities but early fusion across time steps and within each group. Mode \emph{inter-group} extends \emph{group} by replacing the final encoder blocks with fusion blocks that enable cross-group token interactions. Mode \emph{mod} is a special case of \emph{group} with late fusion across all modalities, but early fusion across time steps.}
    \label{fig:fusion_modes}
\end{figure*}

To answer these questions, we explore five different fusion modes, as illustrated in \cref{fig:fusion_modes}:
\begin{itemize}
\item Mode \emph{shared}: Late fusion across modalities and time steps. Parameters are shared across all modalities.
\item Mode \emph{monotemp}: Same as \emph{shared}, but with parameters kept independent for each modality. 
\item Mode \emph{mod}: Late fusion across all modalities but early fusion across time steps. Parameters are kept independent for each modality.
\item Mode \emph{group}: Late fusion across predefined groups of modalities but early fusion across time steps and within groups. Parameters are kept independent across groups.
\item Mode \emph{inter-group}: Similar to \emph{group}, but with the last three encoder blocks replaced by fusion blocks enabling cross-group token interactions. Modalities across groups are subject to intermediate rather than late fusion. 
\end{itemize}

In the last two modes, the goal is to leverage prior knowledge to define sufficiently heterogeneous groups, where sharing encoder parameters and tokens can be beneficial.

\subsection{Pretext Task}
\label{sec:pretext_task}
\paragraph{Multispectral Patch Normalization.} Our tokenizers jointly encode all spectral bands of each modality into shared tokens, implementing \emph{joint-token multispectral fusion}. While this design improves efficiency, we still aim to embed meaningful spectral priors into the reconstruction task. To achieve this, we extend the \emph{patch-wise} target normalization strategy, originally proposed in MAE \cite{he2021} and later adopted in several EO-focused works \cite{wang2022b,hong2024,nedungadi2024,velazquez2025}, with a \emph{patch-group-wise} target normalization  strategy. Specifically, for each modality $m$, we consider partitioning the spectral bands into a set of band groups $g\in\mathcal{G}_m$, and normalizing the targets independently for each patch and each band group.

Using the same notations as previously, let $\mathbf{x}_m, \mathbf{x}^\text{rec}_m \in \mathbb{R}^{I_m / P_m \times I_m / P_m \times D_m \times P^2_m C_m}$ denote the patch-level representations of unnormalized and reconstructed targets for modality $m$, respectively. 
The reconstruction loss with \textbf{patch-group-wise normalization} is: 
\begin{align*}
\hat{\mathcal{L}}^\text{grp} = \frac{1}{\sum_m |\mathcal{P}_m| |\mathcal{T}_m| |\mathcal{G}_m|} \sum_{m,p,t,g} \left \|\hat{\mathbf{x}}^\text{grp}_{m,p,t,g} - \mathbf{x}^\text{rec}_{m,p,t,g}  \right \|_1, \\
\forall m,p,t,g: \quad \hat{\mathbf{x}}^\text{grp}_{m,p,t,g} = \frac{\mathbf{x}_{m,p,t,g} - \mu(\mathbf{x}_{m,p,t,g})}{\sigma(\mathbf{x}_{m,p,t,g})}.
\end{align*}
Our patch-group-wise target normalization generalizes the original patch-wise normalization; specifically, when all bands are grouped into a single set, it reduces to the original strategy used in \cite{he2021,wang2022b,hong2024,nedungadi2024,velazquez2025}.

We find that applying patch-group-wise normalization with well chosen groups $g \in \mathcal{G}_m$, composed of highly correlated bands, yields strong performance. As shown in \cref{sec:ablation}, this target normalization allows joint-token fusion to rival with token-based fusion, while being significantly more computationally efficient.

\paragraph{Masking.} 
We choose a two-stage masking approach: (i) structured masking over modality, spatial, and temporal dimensions, and (ii) unstructured adjustment to meet an overall masking ratio of 75\%. We provide the full details as well as a comparison with previous works in SM \cref{sm:masking}.

\subsection{Downstream Tasks}
\label{sec:downstream_tasks}

\paragraph{Classification/Segmentation Heads.} Following the MAE framework, we transfer only the encoders after SSL pre-training, mitigating domain shift effects related to \texttt{[mask]} tokens \cite{devlin2019,liu2023}.
As a result, the encoder outputs are feature tensors of shape $I_m / P_m \times I_m / P_m \times D_m \times C_e$ for each modality $m$. 

We then attach classification and segmentation heads to process these encoded features (see  \cref{fig:overview}):
\begin{itemize}
\item \emph{Classification heads}: (i) concatenate tokens across all modalities, spatial positions, and time steps, (ii) apply attentive pooling  to aggregate the concatenated tokens \cite{elnouby2024}, and (iii) apply a dense layer with output dimension equal to the number of target classes.
\item \emph{Segmentation heads} first align all modality-specific encoded tensors to a common spatial \emph{token grid of reference}. Then, for each spatial position $p$, they: (i) concatenate tokens across all modalities $m$ and time steps $t$, (ii) apply attentive pooling to aggregate the concatenated tokens, and (iii) apply a dense layer with output dimension equal to the number of target classes.
\end{itemize}

%% file: 4_experiments.tex
\section{Experiments}
\label{sec:experiments}

\subsection{Workflow}
Our workflow varies depending on the evaluated models.

With MAEs, we follow a \emph{three-phase} workflow:
\begin{enumerate}
\item \textbf{Self-supervised pre-training}: The model learns representations from unlabeled data without supervision.
\item \textbf{Probing}:  A task-specific head is trained on top of the frozen pre-trained backbone to assess the quality of the learned representations. Only the task-specific head’s parameters are updated in this phase.
\item \textbf{Fine-tuning}: The entire model–including both the backbone and the task-specific head–is trained end-to-end in a fully supervised manner. 
\end{enumerate}

We consider two distinct settings: \emph{intra-dataset} (SSL pre-training, probing, and fine-tuning on the same dataset) and \emph{cross-dataset} (SSL pre-training on one dataset, probing and fine-tuning on another). In the intra-dataset setting, SSL pre-training is performed exclusively on the union of training and validation sets, with test sets strictly excluded. This ensures that the MAE models do not gain an unfair advantage from any prior exposure to the fine-tuning test domains, \mbox{even indirectly via input modalities.}

With baseline foundation models (FMs) and supervised ViTs, we focus solely on the supervised fine-tuning phase, which is closer to an operational setting.

\subsection{Datasets}
\label{sec:datasets}
We apply our workflows to five medium- to large-scale datasets involving multimodality, multitemporality, and multispectrality:

\begin{itemize}
\item \textbf{TreeSatAI-TS} \cite{ahlswede2023,astruc2024a} – Tree species identification across 15 multi-label classes in Germany, with aerial imagery (RGB + NIR) at 0.2 m resolution, along with Sentinel-1 and 2 time series.

\item \textbf{PASTIS-HD} \cite{garnot2021,astruc2024a} – Agricultural crop segmentation across 19 semantic classes in France, with very high-resolution (VHR) satellite imagery (SPOT 6--7) resampled to 1 m, along with Sentinel-1 and 2 time series.

\item \textbf{FLAIR\#2} \cite{garioud2023,garioud2023b} – Land cover segmentation across 12 semantic classes in France, with aerial and elevation imagery (RGB + NIR + DSM) at 0.2 m resolution, along with Sentinel-2 time series.

\item \textbf{FLAIR-HUB} \cite{garioud2025} – Land cover segmentation across 15 semantic classes in France, with aerial and elevation imagery (RGB + NIR + DEM + DSM) at 0.2 m resolution, along with Sentinel-1 and 2 time series.

\item \textbf{S2-NAIP urban} \cite{s2naip_hf,wolters2023,astruc2024b} – Super-resolution in urban areas of the United States, with aerial imagery at 1.25 m resolution, along with Sentinel-1 and 2 time series.
\end{itemize}

Note that we do not use the full set of labels or input modalities for some datasets (see SM \cref{sm:datasets} for details).

S2-NAIP urban is considered only for SSL pre-training in the \emph{cross-dataset} setting. The remaining four datasets—TreeSatAI-TS, PASTIS-HD, FLAIR\#2, and FLAIR-HUB—are considered for evaluation. 

We also construct filtered versions of the evaluation datasets—derived via Furthest Point Sampling based on geographical coordinates—with 20\% and 5\% of the original samples. These subsets enable us to investigate scaling laws with respect to pre-training and fine-tuning dataset size.

\subsection{Ablation studies}
\label{sec:ablation}

\begin{figure*}[ht]
    \centering
    \includegraphics[width=\linewidth]{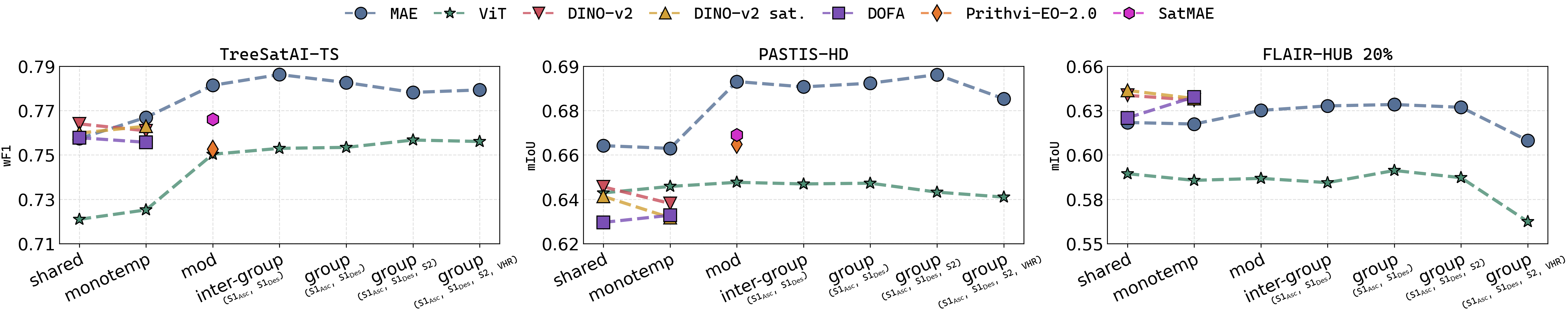}
    \caption{\textbf{Comparison of different multimodal and multitemporal fusion modes for intra-dataset MAE-B models, ViT-B models, and baseline FMs}. We report the weighted F1 score (\%) on TreeSatAI-TS and the mIoU (\%) on PASTIS-HD and FLAIR-HUB 20\%. Results on FLAIR-HUB 20\% with SatMAE and Prithvi-EO-2.0 are markedly low and therefore omitted. Refer to SM \cref{tab:fusion_modes} for exact numbers and additional results with CROMA.}
    \label{fig:fusion_comparison}
\end{figure*}

In this section, we assess the impact of various choices of SSL strategy and model components to guide the design of our final approach. We conduct experiments on TreeSatAI-TS, PASTIS-HD (fold I), and FLAIR-HUB filtered at 20\% (split 1). Our comparison includes intra-dataset MAE and ViT models, alongside several baseline FMs—DINO-v2~\cite{oquab2023}, DINO-v2 sat.~\cite{tolan2024}, DOFA~\cite{xiong2024}, CROMA~\cite{fuller2023}, Prithvi-EO-2.0~\cite{szwarcman2024}, and SatMAE \cite{cong2022}—with hyperparameters carefully set after extensive tuning (see SM \cref{sm:FMs_details}). 

By default, we use the \emph{group} fusion mode, grouping together the Sentinel-1 ascending and descending modalities. For multispectral data, we apply \emph{joint-token} fusion with \emph{patch-group-wise} target normalization during reconstruction, based on spectral band groups exhibiting strong intra-group and weaker inter-group correlations across aerial, Sentinel-1, and Sentinel-2 modalities. Full details are provided in SM~\cref{sm:MAE_MAESTRO,sm:FMs_details}.

\paragraph{Multimodal and Multitemporal Fusion.} We evaluate in \cref{fig:fusion_comparison} and SM \cref{tab:fusion_modes} the different multimodal and multitemporal fusion modes outlined in \cref{fig:fusion_modes}. 

In terms of multimodal fusion, we observe a modest benefit from early fusion among similar modalities, but a significant performance drop when it is applied across dissimilar ones (\emph{group} with all modalities grouped performs worst on all three datasets). Additionally, sharing encoder parameters across modalities did not consistently help when using late fusion (\emph{monotemp} vs \emph{shared}). 

Overall, these results suggest that the potential benefits of synergistic learning across modalities may not outweigh the drawbacks of reduced modality-specific specialization. This finding—likely reflecting the strong heterogeneity of EO modalities—casts doubt on strategies aimed at building universal and modality-agnostic FMs for EO.

Regarding multitemporality, we find that early multitemporal fusion (\emph{mod}, \emph{group}, and \emph{inter-group}) consistently outperforms late multitemporal fusion (\emph{shared} and \emph{monotemp}) by \textbf{+2–3\%} on TreeSatAI-TS (weighted F1) and PASTIS-HD (mIoU), and \textbf{+1\%} on FLAIR-HUB (mIoU). The gains are particularly pronounced for tasks that strongly depend on multitemporal dynamics—such as tree species classification or crop type identification, where temporal signatures and phenology are critical (see SM \cref{sm:impact_multitemporal}). The advantage is also more marked for MAEs than for supervised ViTs, suggesting that effectively exploiting temporal information is both essential and non-trivial.

These findings highlight a potentially overlooked opportunity in multitemporal SSL, which has received less attention than multimodal SSL in prior work. Notably, most existing FMs are inherently monotemporal and thus only compatible with late multitemporal fusion, resulting in a marked performance gap compared to early multitemporal fusion.

\begin{figure}[h]
    \centering
    \includegraphics[width=0.9\linewidth]{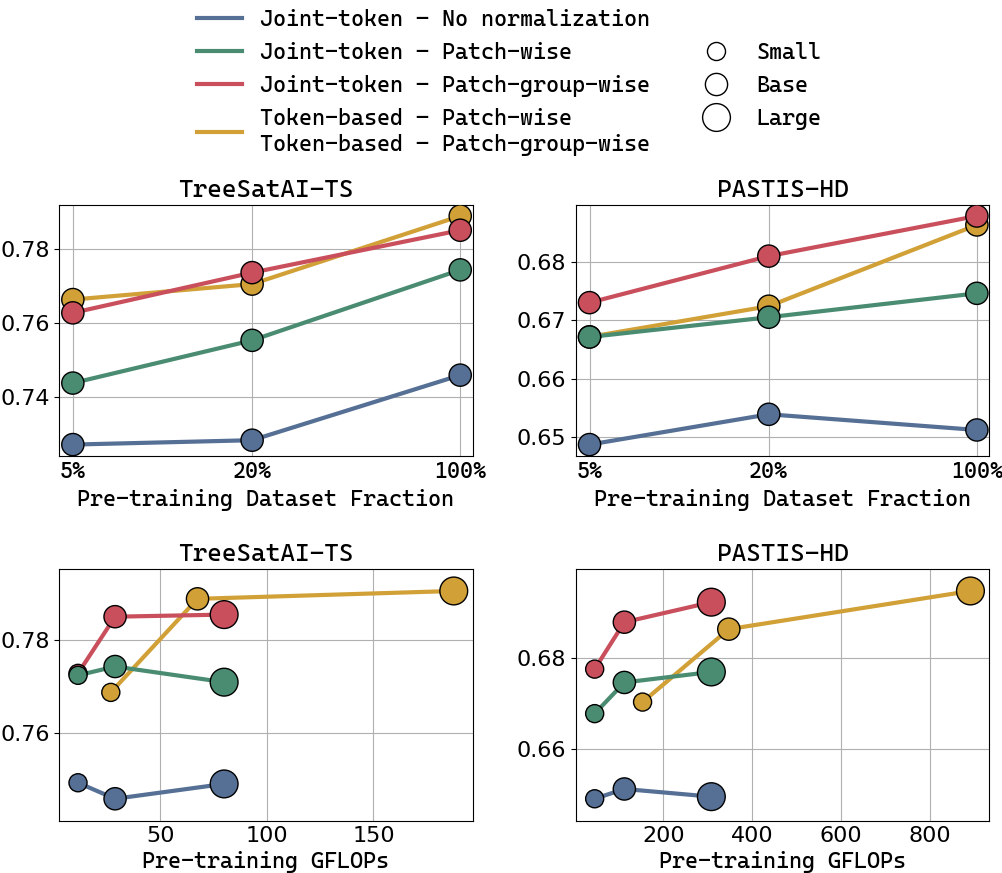}    
    \caption{\textbf{Comparison of different choices of multispectral fusion and target normalization for intra-dataset MAE models.} 
We report the weighted F1 score (\%) on TreeSatAI-TS and the mIoU (\%) on PASTIS-HD, with results shown for varying pre-training dataset fractions (top panels) and varying computational costs (bottom panels). Computational costs are measured as pre-training GFLOPs per forward pass (single batch element) for three model sizes: Small, Base, and Large. Refer to SM \cref{tab:multispectral_fraction,tab:multispectral_size,tab:macs_flops_pretraining} for exact numbers.}
    \label{fig:multispectral_comparison}
\end{figure}

\paragraph{Multispectral Fusion and Target Normalization.} In \cref{fig:multispectral_comparison} and SM \cref{tab:multispectral_fraction,tab:multispectral_size,tab:macs_flops_pretraining}, we evaluate different choices of multispectral fusion and target normalization. Here, we limit our evaluation to TreeSatAI-TS and PASTIS-HD (using fold I) since the performance on FLAIR-HUB is driven mainly by the aerial modality (see SM \cref{sm:impact_modality}), which exhibits only a weak multispectral nature.

We first evaluate the impact of different choices of target normalization within joint-token fusion, where all spectral band groups are combined into the same tokens. Consistent with prior work \cite{he2021,hong2024}, patch-wise normalization improves performance, but our proposed patch-group-wise approach yields significantly better results. %

Strikingly, joint-token fusion with patch-group-wise normalization matches—and sometimes even surpasses—the performance of token-based fusion with patch-wise/patch-group-wise normalization. Note that with token-based fusion, patch-wise and patch-group-wise normalization become equivalent, since each spectral band group forms a separate token \cite{cong2022,irvin2023,tseng2023,hong2024,prexl2024,bountos2025,tseng2025}.

This result is particularly notable given the computational implications: whereas token-based fusion incurs a cost that scales roughly linearly with the number of spectral groups (see SM \cref{sm:flops}), joint-token fusion combined with patch-group-wise normalization achieves similar performance at negligible overhead. This is visible in the performance vs GFLOPs comparison of \cref{fig:multispectral_comparison}.

To our knowledge, the underlying reason for the performance gains with patch-wise normalization remains unclear, even for SSL methods that reconstruct RGB images or videos \cite{he2021,tong2022}. One hypothesis is that it helps balance the pretext task by enforcing uniform difficulty and loss contribution across patches. However, patch-wise normalization may fail to provide such balance for multispectral EO data. For example, if different bands have narrow but non-overlapping histograms, patch-wise normalization tends to reduce to a constant normalization with no patch dependence. Patch-group-wise normalization addresses this by normalizing within both patches and spectral groups.

We hypothesize that this balancing requirement is specific to pixel-space generative SSL. In contrast, latent-space generative SSL methods typically achieve balance implicitly by normalizing either before the loss—through layers such as LayerNorm \cite{assran2023}—or within the loss—through softmax \cite{zhou2021,oquab2023,waldmann2025} or cosine similarity \cite{tseng2025}.

Overall, our findings suggest that, for multispectral data, ensuring a balanced pretext task \cite{oquab2023,vo2024,kerdreux2025,simeoni2025} may be more important than using sophisticated fusion strategies.

\paragraph{Scaling by Dataset Size.}  

\begin{figure}[h]
    \centering
    \includegraphics[width=\linewidth]{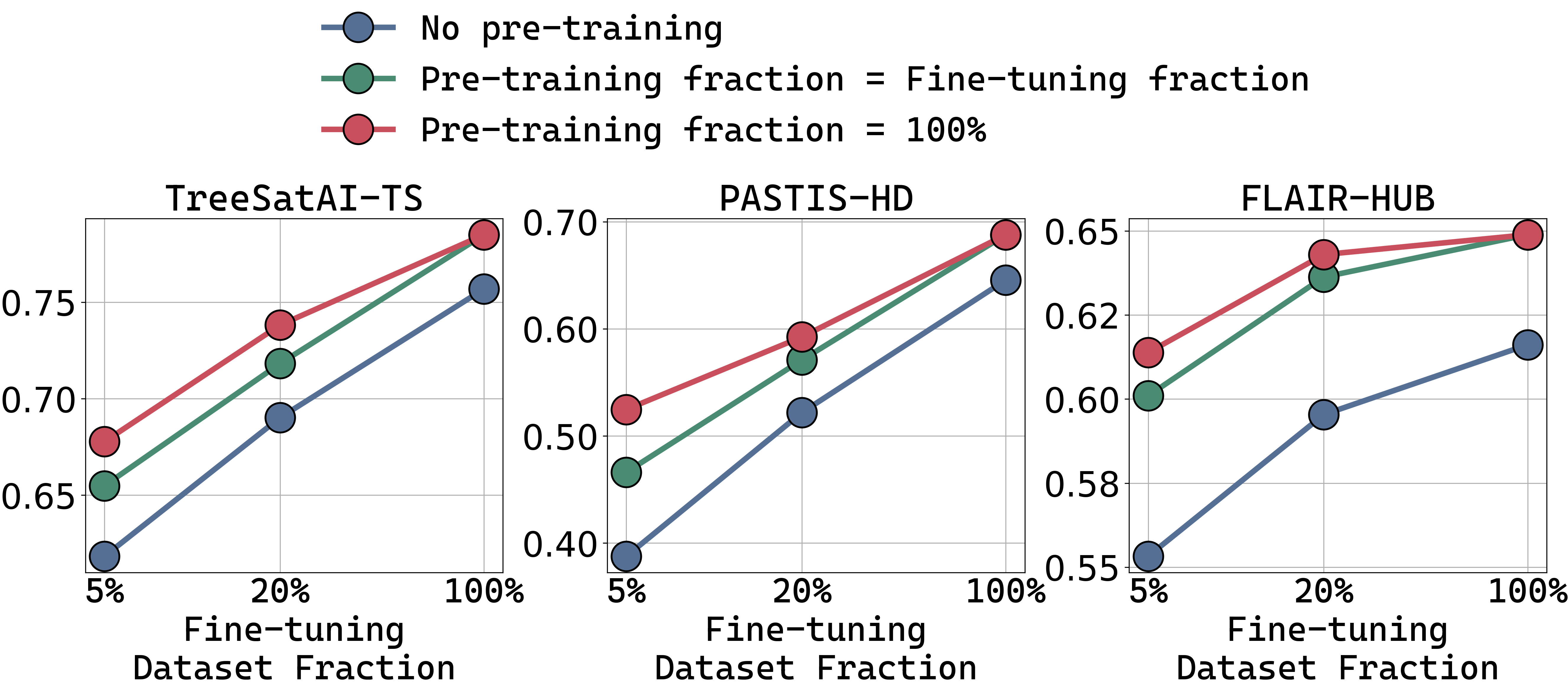}
    \caption{\textbf{Scaling of intra-dataset MAE-B and ViT-B models across pre-training/fine-tuning dataset fractions.} We report the weighted F1 score (\%) on TreeSatAI-TS and the mIoU (\%) on PASTIS-HD and FLAIR-HUB for three fine-tuning dataset fractions: 5\%, 20\%, and 100\%. For each fine-tuning fraction, we compare three pre-training settings: no pre-training, pre-training on the same fraction as fine-tuning, and pre-training on 100\% of the data. Refer to SM \cref{tab:mae_pretrain_finetune} for exact numbers.}
    \label{fig:scaling}
\end{figure}

In \cref{fig:scaling}, we perform an ablation study on the size of pre-training and fine-tuning datasets with intra-dataset MAEs and supervised ViTs (without any pre-training, i.e. with random initialization). MAEs consistently outperforms supervised ViTs, especially when fine-tuning data is limited. Comparing full and reduced pre-training datasets, we see that more unlabeled data systematically improves downstream performance regardless of the amount of labeled fine-tuning data.

\paragraph{Additional Ablation Studies.} In SM, we provide additional ablation studies on the impact of our masking strategy (SM \cref{sm:masking_comparison}), our choices of encodings (SM \cref{sm:ablation_encodings}), the impact of multitemporal components (SM \cref{sm:impact_multitemporal}), and the importance of each modality by dataset (SM \cref{sm:impact_modality}).

\subsection{Design of MAESTRO}
\label{sec:maestro}

Following the analysis in \cref{sec:ablation}, we finalize the choice of SSL strategy and model components for our novel approach. We adopt the \emph{inter-group} fusion mode, grouping together the Sentinel-1 ascending and descending modalities. This choice enables early fusion across time steps and the Sentinel-1 modality pair, while enabling intermediate fusion across other modality pairs. For multispectrality, we use \emph{joint-token} fusion combined with \emph{patch-group-wise} normalization.

We name our approach \textbf{MAESTRO} to reflect how it orchestrates the complex interplay of multimodal, multitemporal, and multispectral components in EO data within the MAE framework, as illustrated in \cref{fig:overview}.

\subsection{Performance}
\label{sec:performance}

We now assess the performance of MAESTRO, alongside that of supervised models and adapted baseline FMs, across the four evaluated datasets. For the adapted baseline FMs, guided by the results of \cref{sec:ablation}, we select the fusion modes \emph{shared} for DINO-v2/DINO-v2 sat./DOFA, \emph{inter-croma} for for CROMA (see SM \cref{sm:adapt_CROMA}), and \emph{mod} for SatMAE and Prithvi-EO-2.0. As noted earlier, these baseline FMs underwent extensive hyperparameter tuning to ensure a fair comparison (see SM \cref{sm:FMs_details}).

\paragraph{Intra-Dataset Evaluation.} As shown in \cref{tab:performance_intra}, MAESTRO constitutes a strong contender in the \emph{intra-dataset} setting. Specifically, it surpasses the previous state-of-the-art (SOTA) by \textbf{+3.8}\% (weighted F1) on TreeSatAI-TS, \textbf{+2.5}\% (mIoU) on PASTIS-HD, and \textbf{+1.5}\% (mIoU) on FLAIR-HUB, while trailing by only -0.8\% (mIoU) on FLAIR\#2 compared to the complex ensemble of \cite{straka2024}.

Intra-dataset evaluations of competing SSL approaches are scarce, largely limited to the TreeSatAI-TS study of \cite{astruc2024a}. MAESTRO’s clear advantage in this setting likely reflects the limitations of other SSL adaptations: (i) suboptimal multitemporal fusion via pixel-median averaging (SatMAE is considered in its non-temporal variant) and (ii) the exclusion of certain modality bands. Compared to AnySat, MAESTRO maintains a strong edge, thanks to its token-based multitemporal fusion and the preservation of fine-grained spatial features throughout the reconstruction task in pixel space \cite{simeoni2025}.

\begin{table}[H]
\centering
\footnotesize
\caption{\textbf{Intra-dataset evaluation of MAESTRO-B, ViT-B, and other supervised baselines and SSL approaches.}. We report the weighted F1 score (\%) on TreeSatAI-TS and the mIoU (\%) on PASTIS-HD, FLAIR\#2, and FLAIR-HUB.
Performance from the two last groups of methods is directly extracted from the literature.
MAESTRO$^\dagger$ models have been pre-trained for twice the number of epochs.}
\label{tab:performance_intra}
\resizebox{0.5\textwidth}{!}{
\addtolength{\tabcolsep}{-0.35em}
\begin{tabular}{ll cccc}
\toprule
Model & \makecell[l]{Fusion\\mode}  
& \makecell{\;\textbf{TSAI-TS} \; \\ \;} 
& \makecell{\textbf{PASTIS-HD} \\ fold I}
& \makecell{\;\textbf{FLAIR\#2} \; \\ split 1} 
& \makecell{\textbf{FLAIR-HUB} \\ split 1} \\
\midrule
\multicolumn{1}{l}{%
  \begin{tabular}{@{}l@{}}
  MAESTRO\\(ours)
  \end{tabular}%
} 
    & inter-group  
    & 78.8\makebox[12pt][l]{\,} 
    & 68.6\makebox[12pt][l]{\,} 
    & 62.6\makebox[12pt][l]{\,}
    & \textbf{65.9}\makebox[12pt][l]{{ \scriptsize\color{Green}$\uparrow$1.6}} \\
\multicolumn{1}{>{\columncolor{black!5}}l}{%
  \begin{tabular}{@{}l@{}}
  MAESTRO$^\dagger$\\(ours)
  \end{tabular}%
}
 & \cellcolor{black!5}inter-group
    & \cellcolor{black!5}\textbf{79.4}\makebox[12pt][l]{{ \scriptsize\color{Green}$\uparrow$3.8}}
    & \cellcolor{black!5}\textbf{69.0}\makebox[12pt][l]{{ \scriptsize\color{Green}$\uparrow$2.5}}
    & \cellcolor{black!5}\textbf{63.3}\makebox[12pt][l]{{ \scriptsize\color{Red}$\downarrow$0.8}}
    & \cellcolor{black!5}65.8\makebox[12pt][l]{\,} \\
 \greyrule
ViT & inter-group & 75.6\hspace{12pt} & 64.5\hspace{12pt} & 58.2\hspace{12pt} & 62.1\hspace{12pt} \\
\greyrule
TSViT \cite{tarasiou2023vits} &  & -\hspace{12pt} & 65.4\hspace{12pt} & -\hspace{12pt} & -\hspace{12pt} \\
\rowcolor{black!5} UTAE-MM \cite{garnot2022} &  & -\hspace{12pt} & 66.3\hspace{12pt} & -\hspace{12pt} & -\hspace{12pt} \\
UT\&T \cite{garioud2023} &  & 56.7\hspace{12pt} & -\hspace{12pt} & 56.9\hspace{12pt} & -\hspace{12pt} \\
\rowcolor{black!5} Ens-04 \cite{straka2024}&  & -\hspace{12pt} & -\hspace{12pt} & \textbf{64.1}\hspace{12pt} & -\hspace{12pt} \\
UPerFuse \cite{garioud2025} & & -\hspace{12pt} & -\hspace{12pt} & -\hspace{12pt} & \textbf{64.3}\hspace{12pt} \\
\greyrule
                   CROMA \cite{fuller2023} & & 61.0\hspace{12pt} & -\hspace{12pt} & -\hspace{12pt} & -\hspace{12pt} \\
\rowcolor{black!5} SatMAE \cite{cong2022} &  & 61.5\hspace{12pt} & -\hspace{12pt} & -\hspace{12pt} & -\hspace{12pt} \\
                   ScaleMAE \cite{reed2023} &  & 62.5\hspace{12pt} & -\hspace{12pt} & -\hspace{12pt} & -\hspace{12pt} \\
\rowcolor{black!5} OmniSat \cite{astruc2024a} & & 74.2\hspace{12pt} & -\hspace{12pt} & -\hspace{12pt} & -\hspace{12pt}  \\
                   AnySat \cite{astruc2024b} & & \textbf{75.1}\hspace{12pt} & \textbf{66.5}\hspace{12pt} & 55.1\hspace{12pt} &  -\hspace{12pt} \\  
\bottomrule
\end{tabular}
}
\end{table}

\input{tables/tab_Perf_FM}

\paragraph{Cross-Dataset Evaluation.} 
In \cref{tab:performance_cross}, we evaluate MAESTRO in the \emph{cross-dataset} setting, with models pre-trained on the large-scale FLAIR-HUB and S2-NAIP urban datasets, following the analysis in \cref{sec:ablation}. MAESTRO achieves SOTA results on TreeSatAI-TS (\textbf{+2.7}\% weighted F1) and PASTIS-HD (\textbf{+1.4}\% mIoU) by outperforming the strongest adapted baselines, reflecting the limitations of late multitemporal fusion in DINO-v2, DINO-v2 sat., CROMA, and DOFA, as well as domain shifts in temporal and spectral configurations in Prithvi-EO 2.0 and SatMAE.

However, this advantage does not extend to FLAIR\#2 and FLAIR-HUB, where MAESTRO pre-trained on S2-NAIP roughly matches DOFA and midly trails DINO-v2/DINO-v2 sat.. This likely stems from these datasets relying mainly on monotemporal aerial imagery (see SM \cref{sm:impact_modality}), reducing the influence of multitemporal fusion. Still, MAESTRO performs well given (i) the domain shifts between S2-NAIP urban and FLAIR, and (ii) the fact that DOFA and DINO-v2/DINO-v2 sat. were pre-trained on much larger and potentially better balanced datasets \cite{oquab2023}.

\Cref{tab:performance_cross} also shows that adaptation is crucial: our \emph{adapted} FMs consistently outperform \emph{original} FMs. It is noteworthy that even after adaptation, FMs tend to perform best when the dominant fine-tuning modality closely matches the one used during pretraining (see SM \cref{tab:baseline_details}). As supported by \cref{sec:ablation}, this might reflect the high heterogeneity of EO data—across modalities, resolutions, spectral bands, and axes structures—making it challenging to design universal FMs that perform well across diverse downstream tasks.

%% file: tables/tab_Perf_FM.tex
\begin{table*}[!t]
\centering
\footnotesize
\caption{\textbf{Cross-dataset evaluation of MAESTRO, adapted baseline FMs, and original baseline FMs.}
We report weighted F1 (\%) on TreeSatAI-TS and mIoU (\%) on PASTIS-HD, FLAIR\#2, and FLAIR-HUB.
\mono~denotes support only for monotemporal inputs, while \multitemp~denotes support for multitemporal inputs.
\eyeslash~indicates that the original model discards some modality bands.
For a fair comparison, we adapted baseline FMs to handle an arbitrary number of dates (\greenmultitemp) and/or to retain all modality bands (\greeneye). Results for the original FMs are directly extracted from the literature.}
\label{tab:performance_cross}
\resizebox{\textwidth}{!}{
\addtolength{\tabcolsep}{-0.25em}
\begin{tabular}{lllr ccccc cccc}
\toprule
\multirow{2.6}{*}{Model}  & \multirow{2.6}{*}{\makecell[c]{Fusion\\mode}}  & \multicolumn{2}{c}{Pre-training dataset}  & \multicolumn{4}{c}{Input modalities}  
& \multirow{2.6}{*}{\makecell[c]{\textbf{TSAI-TS}\\ \;}}
& \multirow{2.6}{*}{\makecell[c]{\textbf{PASTIS-HD}\\fold I}}
& \multirow{2.6}{*}{\makecell[c]{\textbf{FLAIR\#2} \\split I}}
& \multirow{2.6}{*}{\makecell[c]{\textbf{FLAIR-HUB}\\split I}} \\
\cmidrule(lr){3-4} \cmidrule(lr){5-8}
 &  & Name & GPixels & S1
& S2 & VHR  & DEM & & 
&   
&  \\
\midrule
MAESTRO (ours) & inter-group & FLAIR-HUB & 63 &\multitemp & \multitemp & \yes & \yes & \textbf{79.6} & \textbf{68.0}  & - & - \\
\rowcolor{black!5} MAESTRO (ours) & inter-group & S2-NAIP urban & 64 & \multitemp & \multitemp & \yes & \no & 78.8 & 67.4 & 62.6 & 64.6 \\
\midrule
\multicolumn{12}{c}{\textbf{Adapted FMs}} \\
\greyrule
                    DINO-v2 \cite{oquab2023}            & shared & LVD-142M & 24,592 & \no & \greenmultitemp~\greeneye & \greeneye & \no & 76.7 & 64.4 & \textbf{64.2} & 66.0 \\
\rowcolor{black!5}  DINO-v2 sat. \cite{tolan2024}  & shared &  Maxar Vivid2 & 1,179 & \no & \greenmultitemp~\greeneye & \greeneye & \no & 76.3 & 64.0 & 63.5 & \textbf{66.0} \\
                    DOFA \cite{xiong2024}               & shared & DOFA MM & 2,115  & \greenmultitemp & \greenmultitemp~\phantomeye & \yes & \no & 76.0 & 62.9 & 62.3 & 65.1 \\
\rowcolor{black!5}  CROMA \cite{fuller2023}             & inter-croma& SSL4EO & 139 & \greenmultitemp & \greenmultitemp~\phantomeye &  \no & \no & 70.5 & 65.0 & 39.0 & 44.3 \\
\rowcolor{black!5}  Prithvi-EO-2.0 \cite{szwarcman2024}  & mod & HLS& 1,113 & \no & \greenmultitemp~\greeneye & \no & \no & 75.6 & 66.2 & 41.8 & 44.9  \\
                    SatMAE \cite{cong2022}    & mod &  fMoW RGB+S &  54 & \no & \greenmultitemp~\greeneye & \no & \no & \textbf{76.9} & \textbf{66.6}  &  42.5 & 45.0 \\
\midrule
\multicolumn{12}{c}{\textbf{Original FMs}} \\  
\greyrule
DINO-v2 GeoSA \cite{luo2025} &- &  LVD-142M  & 24,592 & \no & \no & \eyeslash & \no & - & - & 63.5 & -  \\
\rowcolor{black!5} DOFA \cite{xiong2024}  & - & DOFA MM & 2,115  & \mono & \mono & \yes & \no & 71.6 & - & - & - \\
                 Terramind \cite{jakubik2025} & -& TerraMesh & 4,390 & \mono  & \mono  & \no  & \yes & - & 43.1 & - & -  \\ 
FoMo-Net \cite{bountos2025} & - &  FoMo-Bench & 395 & \multitemp & \multitemp &  \yes & \yes & 45.0 & - & 47.8 & -  \\ %
\rowcolor{black!5} Galileo \cite{tseng2025} & - & Galileo  & 28 & \multitemp & \multitemp & \no & \yes & - & 39.2 & - & - \\
\bottomrule
\end{tabular}
}
\end{table*}

%% file: 5_conclusion.tex
\section{Conclusion}

In this work, we explored how to adapt SSL approaches to the distinctive characteristics of EO data, focusing on three axes of heterogeneity: multimodality, multitemporality, and multispectrality.

(i) \textbf{Multimodality} – Early fusion offers slight benefits when modalities are similar but can hurt performance when they differ significantly, raising doubts about the viability of universal, modality-agnostic EO foundation models.  (ii) \textbf{Multitemporality} – Early fusion consistently outperforms late fusion for tasks that strongly rely on multitemporal dynamics, revealing an underexplored opportunity in multitemporal SSL. Most current foundation models are inherently monotemporal, requiring late multitemporal fusion at fine-tuning, which can incur significant performance drops. (iii) \textbf{Multispectrality} – Joint-token early fusion combined with a proper target normalization for reconstruction matches the performance of token-based fusion while being significantly more efficient.

Building on these findings, we proposed \textbf{MAESTRO}—an adaptation of the MAE that effectively orchestrates the use of multimodal, multitemporal, and multispectral data in EO. Evaluated on four benchmarks, MAESTRO achieves SOTA results on tasks with strong multitemporal components, while remaining competitive on others.

We hope that our work will contribute to the design of SSL strategies that specifically account for the distinctive characteristics of EO data.

%% file: 6_experimental_details.tex
\clearpage
\maketitlesupplementarysingle

The appendix is structured as follows: 
\begin{itemize}
\item \cref{sm:comparison_high_level} provides a high-level comparison of MAESTRO and related approaches from the litterature;
\item \cref{sm:sec_exp_details} provides our full experimental details; 
\item \cref{sm:detailed_results} reports the detailed results for the experiments presented in \cref{fig:fusion_comparison,fig:multispectral_comparison,fig:scaling}; 
\item \cref{sm:add_results} reports additional ablation studies; 
\item \cref{sm:quali_results} offers a qualitative analysis on inference results; 
\item \cref{sm:flops} reports computational costs in FLOPs.
\end{itemize}

\section{Comparison Table of MAESTRO versus Prior Work}
\label{sm:comparison_high_level}

We provide in \cref{tab:comparison_high_level} a high-level overview of the differences between our approach, MAESTRO, and prior SSL work in Earth observation. 

\begin{table*}[ht!]
\centering
\footnotesize
\caption{\textbf{Comparison of MAESTRO with prior SSL work in EO}. We indicate the characteristics of the original versions of the related SSL approaches (i.e. before the adaptation of some of these methods presented in \cref{sm:FMs_details}). $\oplus$ means exclusive OR; \share means parameter sharing; \jointt means joint-token, \tbased token-based; \early means early fusion, \inter intermediate fusion.
}
\label{tab:comparison_high_level}
\begin{tabular}{clllllll}
\toprule
 & Model &  Multimodal & Multitemporal & Multispectral & VHR & DEM/DSM & Remark  \\
\midrule
\multirow{17}{*}{\makecell{MAE/SimMIM\\style}}
 & ScaleMAE \cite{reed2023} &  \no & \no & \no & \yes & \no & - \\
 & Cross-Scale MAE \cite{tang2023} &  \no & \no & \no & \yes & \no & - \\
 & SatMAE \cite{cong2022} & \no & $\oplus$\yes~\tbased\early & $\oplus$\yes~\tbased\early & \no & \no & Limited to 3 dates  \\
 & Prithvi v1/v2 \cite{jakubik2023,szwarcman2024} & \no & \yes~\tbased\early & \yes~\jointt\early & \no & \no & Limited to 3--4 dates (v1--v2)  \\ 
 & CROMA \cite{fuller2023}  & \yes~\tbased\inter & \no & \yes~\jointt\early & \no & \no & - \\
 & RingMo \cite{sun2023} & \yes~\share & \no & \no & \yes & \no & Frequency-enhanced MIM \\
 & USat \cite{irvin2023} & \yes~\tbased\early & \no & \yes~\tbased\early & \yes & \no & - \\
 & Billion-scale FM \cite{cha2023} & \yes~\share & \no & \no & \yes & \no & - \\
 & GFM \cite{mendieta2023} & \yes~\share & \no & \no & \yes & \no & - \\
 & msGFM \cite{han2024}  & \yes~\share & \no & \yes~\jointt\early & \yes & \yes & - \\
 & MMEarth \cite{nedungadi2024} & \yes~\share & \no & \yes~\jointt\early & \no & \yes &  ConvNeXt V2 encoder \\
 & SenPa-MAE \cite{prexl2024} & \yes~\share & \no & \yes~\tbased\early & \no & \no & - \\ 
 & SeaMo \cite{li2024b}& \yes~\tbased\early & \yes~\tbased\inter & \yes~\jointt\early & \no & \no & Limited to 4 dates  \\
 & DOFA \cite{xiong2024} & \yes~\share & \no  & \yes~\jointt\early & \yes & \no & - \\
 & EarthMAE \cite{velazquez2025} & \yes~\tbased\early & \yes~\tbased\early & \yes~\jointt\early & \yes & \yes & - \\
 & FoMo-Net \cite{bountos2025} & \yes~\tbased\early & \yes~\tbased\early & \yes~\tbased\early & \yes & \yes & - \\
 & Copernicus-FM  \cite{wang2025a} & \yes~\share & \no & \yes~\jointt\early & \no & \yes & - \\
 \midrule
\multirow{10}{*}{\makecell{Other\\Approaches}}
 & CMID \cite{muhtar2023}&  \no & \no & \no & \yes & \no & Mix of contrastive and MIM \\
 & DINO-v2 sat. \cite{tolan2024} & \no & \no  & \no & \yes & \no & - \\
 & DINO-v2 GeoSA \cite{luo2025}& \no & \no  & \no & \yes & \no & - \\
 & OmniSat \cite{astruc2024a} & \yes~\tbased\inter  & \yes~\jointt\early & \yes~\jointt\early & \yes & \yes & - \\ 
 & AnySat \cite{astruc2024b} & \yes~\tbased\inter  & \yes~\jointt\early  & \yes~\jointt\early & \yes &  \yes & JEPA-style \\
 & SkySense v1/v2 \cite{guo2024,zhang2025}&  \yes~\tbased\inter & \yes~\tbased\inter & \yes~\jointt\early & \yes & \no & - \\
 & SkySense++ \cite{wu2025}& \yes~\tbased\inter & \yes~\tbased\inter & \yes~\jointt\early & \yes & \no & Early parameter sharing \\
 & Presto \cite{tseng2023} & \yes~\tbased\early & \yes~\tbased\early & \yes~\tbased\early & \no & \yes & Pixel level w/o spatial context \\
 & Galileo \cite{tseng2025}&  \yes~\tbased\early  & \yes~\tbased\early & \yes~\tbased\early  & \no & \yes & Mix of MAE and JEPA  \\
 & Terramind \cite{jakubik2025} & \yes~\tbased\early & \no & \yes~\jointt\early & \no & \yes & - \\
\midrule
\multirow{1}{*}{Ours}
 & MAESTRO & \yes~\tbased\inter & \yes~\tbased\early & \yes~\jointt\early & \yes & \yes & - \\
\bottomrule
\end{tabular}
\end{table*}

\section{Experimental Details}
\label{sm:sec_exp_details}

In this section, we provide our full experimental details. We provide details on the four evaluated datasets (\cref{sm:datasets}), details shared across models (\cref{sm:shared_across_models}), details specific to MAEs/MAESTRO (\cref{sm:MAE_MAESTRO}), and details specific to the adapted baseline FMs (\cref{sm:FMs_details}). Additionally, we provide tables reporting the exhaustive list of hyperparameter values (\cref{sm:hyperparam_tables}).

\subsection{Experimental Details on the Datasets}
\label{sm:datasets}

In this subsection, we provide additional details on the choice of input modalities and labels for the four evaluated datasets (\cref{sm:datasets_tsai,sm:datasets_pastis,sm:datasets_flair2,sm:datasets_flair_hub}) and the additional pre-training dataset S2-NAIP urban (\cref{sm:datasets_s2naip}). We also report further details on the raw preprocessing steps applied to some input modalities of these datasets (\cref{sm:datasets_preproc}).

\subsubsection{TreeSatAI-TS}
\label{sm:datasets_tsai}
TreeSat was introduced in \cite{ahlswede2023} and comprises 50,381 tiles of 60 m $\times$ 60 m in Germany, featuring aerial imagery (RGB + NIR) at 0.2 m resolution, together with monotemporal Sentinel-1 and Sentinel-2 data (Sentinel-2 provided as seasonal medians). It was later extended as TreeSatAI-TS in \cite{astruc2024a}, adding Sentinel-1 and Sentinel-2 time series covering the full year closest to the aerial acquisition. In this work, we retain four distinct modalities: aerial imagery, Sentinel-1 time series in both orbits (ascending and descending), and Sentinel-2 time series. We disregard the original monotemporal Sentinel-1 and Sentinel-2 data.

The original labels introduced in \cite{ahlswede2023} were formulated as regression targets, representing the spatial fraction of each tree species within a patch. Following \cite{astruc2024a,astruc2024b}, we recast this as a multi-label classification task: a species class is considered present if its spatial fraction exceeds 0.07.

\subsubsection{PASTIS-HD}
\label{sm:datasets_pastis}
PASTIS was introduced in \cite{garnot2021} and comprises 433 tiles of 1280 m $\times$ 1280 m in France, featuring Sentinel-2 time series spanning approximately one year. It was later extended as PASTIS-R in \cite{garnot2022} by adding Sentinel-1 time series covering about 70 dates in both ascending and descending orbits, and further expanded in \cite{astruc2024a} as PASTIS-HD with SPOT 6–7 imagery resampled to 1 m resolution. We retain four distinct modalities: SPOT 6–7 imagery, Sentinel-1 time series in both orbits (ascending and descending), and Sentinel-2 time series.

The crop segmentation labels in \cite{garnot2021} covered both semantic and panoptic segmentation. Here, we retain only the semantic segmentation labels, omitting panoptic segmentation. Restricting to semantic segmentation allows us to incorporate a spatialized fine-tuning task without introducing additional complexities related to specialized heads or loss functions. We assume that the spatialized nature of the task (e.g., semantic segmentation vs. classification) plays a more decisive role than the precise segmentation type (e.g., panoptic vs. semantic) when benchmarking different SSL approaches.

\subsubsection{FLAIR\#2}
\label{sm:datasets_flair2}
FLAIR was introduced in \cite{garioud2022} and comprises 77,762 tiles in France, with aerial and elevation imagery (RGB + NIR + DSM) at 0.2 m resolution. It was extended as FLAIR\#2 in \cite{garioud2023,garioud2023b} with Sentinel-2 time series spanning a full year in the form of superpatches, covering a larger spatial extent than the aerial imagery (400 m $\times$ 400 m vs. 102.4 m $\times$ 102.4 m) to provide additional spatial context. Following \cite{astruc2024b}, we crop the Sentinel-2 time series to match the extent of the VHR imagery, discarding 93.5\% of pixels. We also include elevation imagery at 0.2 m resolution, extracted from the FLAIR-HUB extension \cite{garioud2025} on the 77,762 FLAIR\#2 tiles. In total, we retain three distinct modalities: aerial imagery (RGB + NIR), elevation imagery (DEM + DSM), and Sentinel-2 time series.

We use the land cover semantic segmentation labels following the filtering procedure described in \cite{garioud2022,garioud2023,garioud2023b}: we retain only 12 of the 18 original classes, excluding the remaining 6 classes from the loss and metric computations.

\subsubsection{FLAIR-HUB} 
\label{sm:datasets_flair_hub}
FLAIR-HUB \cite{garioud2025} is an extension of FLAIR\#2 with 241,100 tiles of 102.4 m $\times$ 102.4 m. Compared to FLAIR\#2, FLAIR-HUB enriches the elevation imagery (DEM + DSM) and treats it as a separate modality, crops the Sentinel-2 imagery to the extent of the aerial imagery (as we do in our version of FLAIR\#2), and adds Sentinel-1 time series spanning a full year, as well as SPOT 6–7 imagery and historical aerial imagery. We retain five distinct modalities: aerial imagery (RGB + NIR), elevation imagery (DEM + DSM), Sentinel-1 time series in both orbits, and Sentinel-2 time series. We disregard the SPOT 6–7 imagery, which did not significantly improve fine-tuning performance—likely due to redundancy with the aerial modality—and the historical imagery, due to its lack of synchronicity with the labels. 

We use the land cover semantic segmentation labels following the filtering procedure in \cite{garioud2025}: we retain 15 of the 18 original classes, excluding the remaining 3 from the loss and metric computations. We omit the crop semantic segmentation labels.

\subsubsection{S2-NAIP urban} 
\label{sm:datasets_s2naip}

S2-NAIP \cite{s2naip_hf,wolters2023} covers the continental United States, featuring National Agriculture Imagery Program (NAIP) aerial imagery at 1.25 m resolution, together with Sentinel-1, Sentinel-2, and Landsat time series at 10 m resolution.

We construct the urban subset following \cite{astruc2024b}, i.e. we restrict the original S2-NAIP footprints to those overlapping with the urban set defined in \cite{wolters2023}, which includes only locations within a 5 km radius of U.S. cities with populations of at least 50,000. This filtering ensures that the dataset remains balanced and avoids being dominated by homogeneous rural landscapes \cite{mendieta2023,wolters2023}.

The resulting subset contains 167,397 tiles of size 640 m $\times$ 640 m, covering a total area of 68,565 $\text{km}^2$.

We retain three distinct modalities: NAIP imagery, Sentinel-1 time series from both ascending and descending orbits, and Sentinel-2 time series. 

We exclude the OpenStreetMap and WorldCover labels, as S2-NAIP urban is used exclusively for MAESTRO pre-training in the cross-dataset evaluation of \cref{sec:performance}.

\subsubsection{Raw Preprocessing}
\label{sm:datasets_preproc}
In TreeSatAI-TS, FLAIR\#2, and FLAIR-HUB, the Sentinel-1 backscattering coefficient data is provided in linear scale. We apply a logarithmic transformation to express it in decibel (dB) scale (up to a multiplicative constant). For PASTIS-HD, the Sentinel-1 data is already in dB scale; however, we retain only the first two channels, corresponding to vertical (VV) and horizontal (VH) polarizations, and discard the third channel expressing their ratio (VV/VH).

In S2-NAIP urban, we retain only ten out of the twelve Sentinel-2 bands, keeping those at 10 m and 20 m resolutions but excluding the two available only at 60 m resolution.

In FLAIR\#2 and FLAIR-HUB, we also apply a simple preprocessing step to the elevation imagery (DEM + DSM). We recast it as the DEM and a rescaled elevation defined as $10^{3} \times (\text{DSM}-\text{DEM})$, ensuring that both channels have comparable value ranges.

The TreeSatAI-TS aerial imagery tiles have a slightly larger spatial extent than the other modalities (304 $\times$ 304 pixels, corresponding to 60.8 m, instead of 300 $\times$ 300 pixels). We therefore apply a centered crop to align them spatially with the other inputs.

\subsection{Experimental Details Shared across Models}
\label{sm:shared_across_models}

In this subsection, we provide details on the preprocessing pipeline (\cref{sm:preprocessing}), positional/temporal encodings (\cref{sm:encodings}), data augmentation (\cref{sm:data_aug}), regularization (\cref{sm:regu}) and optimizers (\cref{sm:optimizer}).

\subsubsection{Preprocessing}
\label{sm:preprocessing}
Here we provide the full details on the preprocessing pipeline introduced in \cref{sec:architecture}. The full implementation is available in \verb|ssl_models/dataset/dataset.py| in our code.

\paragraph{Dataset-specific crop hyperparameter}  
Each dataset $\mathcal{D}$ has a specific hyperparameter controlling the \emph{spatial extent of crops} within original tiles. For classification tasks, where labels apply to entire tiles, we set the crop extent equal to the original tile extent. For segmentation tasks, the crop extent can be smaller, with the same crop applied to the segmentation labels. Choosing the crop size involves balancing two competing effects:
\begin{itemize}
\item Smaller crops increase the spatial resolution of the token grid under a fixed token budget. For a fixed token budget $(I_m / P_m)^2 D_m$, reducing the image size $I_m$ reduces the patch size $P_m$, mitigating the risk of an information bottleneck in the tokenizer (the model’s entry block).
\item Larger crops provide more spatial context for the model.
\end{itemize}

When cropping to a smaller extent than the original tile, we apply for each epoch a repetition factor $R_\mathcal{D}$ equal to the ratio of the original tile area to the cropped area. This ensures that the total spatial area seen by the model in each epoch matches that of the full dataset. Practically, this means the dataset length is scaled by $R_\mathcal{D}$: each tile, originally mapped to a single index, is now associated with $R_\mathcal{D}$ indices, effectively repeating it that many times with different crops.

The \emph{crop sampling} strategy differs between training and validation/testing:
\begin{itemize}
\item \emph{Training:} the crop is sampled randomly from all valid crops within the tile, subject only to the constraint that crop boundaries align with integer pixel indices for each modality $m$.
\item \emph{Validation and testing:} tiles are partitioned into non-overlapping crops, which are iterated over exhaustively. The same repetition factor is applied in the test epoch, ensuring test metrics cover the exact same spatial footprint as the original tiles.
\end{itemize}

In practice, we use the full tile extent for TreeSatAI-TS, FLAIR\#2, and FLAIR-HUB. For PASTIS-HD and S2-NAIP urban, however, we found it beneficial to substantially reduce the crop size. This allows us to set the number of temporal bins as $D_m = 16$ for Sentinel-2 and $D_m = 4$ for Sentinel-1, while setting the patch size as $P_m = 2$ for both modalities. Without cropping, keeping the same token budget and number of temporal bins would have required to set $P_m = 16$ for these modalities.

\paragraph{Modality-specific hyperparameters.}  
Each modality $m$ in the dataset $\mathcal{D}$ has a configurable set of preprocessing hyperparameters:
\begin{itemize}
\item The \emph{image size} $I_m$, which by default corresponds to the number of pixels covering the spatial extent of the crop (however, this can be adjusted for cross-dataset MAESTRO and baseline FMs to yield a given token grid size while retaining a specific transferred patch size $P_m$);
\item The \emph{target number of temporal bins} $D_m$;
\item Whether a \emph{snow/cloud mask} mask is used to guide valid time-step selection within temporal bins, and, if so, the probability threshold above which values are disregarded;
\item A \emph{constant multiplicative normalization factor} applied to the modality.
\end{itemize}

\paragraph{Preprocessing steps.}  
The \verb|__getitem__| method of our datasets operates in three steps:
\begin{enumerate}[label=(\roman*)]
\item Determine the sampled dataset tile based on the dataset index;
\item Sample a spatial crop shared across all modalities $m \in \mathcal{D}$;
\item Process each modality $m$ based on the sampled crop.
\end{enumerate}

Processing step (iii) for each modality $m$ proceeds as follows:
\begin{enumerate}[label=(\alph*)]
\item If the number of original time steps $T_m$ is not a multiple of $D_m$, randomly truncate the sequence to $D_m \times \left\lceil T_m / D_m \right\rceil$ consecutive time steps.
\item Read the corresponding tile section defined by the crop’s spatial extent and the (potentially truncated) temporal extent. To minimize I/O overhead, we use lazy loading:
\begin{itemize}
\item For \verb|.tif| files: read with \verb|rasterio| using window reading.
\item For \verb|.npy| files: read with \verb|numpy.load| using the argument \verb|mmap_mode="r"|.
\item For \verb|.h5| files: read with \verb|h5py.File| in read mode, and index the required array sections.
\end{itemize}
\item Reshape the array into $D_m$ temporal bins.
\item For each temporal bin, sample one time step inside the bin:
\begin{itemize}
\item If a snow/cloud mask is used, filter out time steps that include any mask values above the configured threshold. An exception occurs when no valid time steps remain, in which case all time steps in the bin are retained.
\item Select a single time step from the remaining ones. During training, the selection is random to serve as data augmentation. During validation and testing, we maximize representativeness by: computing the pixel-wise median across valid time steps in the temporal bin; computing the mean absolute deviation to this median for each time step; and selecting the time step with the lowest mean absolute deviation.
\end{itemize}
\item If the image size $I_m$ does not match the number of pixels covering the spatial extent of the crop, reinterpolate the array spatially with \verb|torch.nn.functional.interpolate| using \verb|mode="nearest"|.  
\item Scale the array by the configured multiplicative normalization factor.
\end{enumerate}

\subsubsection{Positional/Temporal Encodings}
\label{sm:encodings}

As noted in ~\cref{sec:architecture}, we do not include explicit modality encodings, since we use modality-specific tokenizers and learnable modality-specific \verb|[mask]| tokens that implicitly encode the source or target modality for each token. However, we do include spatial and temporal positional encodings, as detailed below.

\paragraph{Spatial encodings.} To encode spatial information, we follow the common practice of using two-dimensional sine–cosine positional encodings \cite{he2021,vaswani2017,touvron2021}. As in Scale-MAE \cite{reed2023}, we scale these positional encodings according to the ground sampling distance of each modality. Concretely, we compute the \emph{least common multiple} (LCM) of the token grid sizes across modalities, i.e., $\lcm_{m \in \mathcal{D}} \{ I_m / P_m \}$, and generate positional encodings on the high-resolution grid defined by this LCM. For each modality $m$, we then obtain its positional encodings by downsampling the high-resolution grid, averaging over all high-resolution grid cells corresponding to each low-resolution grid cell.

\paragraph{Temporal encodings.} To encode temporal information, we first extract the day of year and hour of day for the selected time step in each temporal bin (see~\cref{sm:preprocessing}). The day of year is normalized by $365.25$ and the hour of day by $24$. Applying sine and cosine transformations to these normalized values yields four temporal features per time step.

In addition, for each tile, we define a reference date shared across all modalities~$m$ to capture temporal differences across years, rather than only within a single year (as we would be limited to with just sine/cosine encodings). For each time step, we compute the difference between its acquisition date and this reference date, obtaining a fifth temporal feature. This value is then duplicated four times, resulting in a total of eight temporal features per time step.

\paragraph{Aggregation.} 
Spatial and temporal encodings are aggregated by concatenation. In the encoder with latent dimension $C_e$, the spatial encodings occupy $C_e-8$ dimensions, while the temporal encodings use $8$ dimensions, leading to $C_e$ dimensions after concatenation. The same approach is applied in the decoder with latent dimension $C_d$.

\subsubsection{Data Augmentation}
\label{sm:data_aug}

We use up to three types of data augmentation:
\begin{itemize}
\item \emph{Random spatial cropping}: During training, a crop of the configured extent is randomly sampled within the original tile at the start of preprocessing. This is disabled for validation and testing, where tiles are instead partitioned into non-overlapping crops processed exhaustively in each epoch. When the crop extent matches the original tile extent, this augmentation is implicitly disabled, even during training.
\item \emph{Random time step selection}: During training, time steps are randomly sampled from the valid steps within each temporal bin. For validation and testing, we instead select time steps that maximize representativeness among valid steps.
\item \emph{D4 augmentation}: Throughout training, validation, and testing, synchronized D4 transformations are applied across all modalities and semantic segmenation labels just after the preprocessing pipeline. We retain this augmentation for validation and testing, as the model is assumed to have learned equivariance to these transformations during training.
\end{itemize}

\subsubsection{Regularization}
\label{sm:regu}
As regularization, we apply an Exponential Moving Average (EMA) of the model weights during fine-tuning \cite{morales2024}, similar to Stochastic Weight Averaging \cite{izmailov2018}. Concretely, an EMA of the weights is updated at each epoch with a smoothing window equal to 20\% of the total fine-tuning epochs. Denoting the model weights by $\theta$ and their EMA by $\theta_\text{EMA}$, the EMA weights are initialized as $\theta$ and updated at each epoch as:
\begin{align*}
\theta_\text{EMA} & = \alpha \theta_\text{EMA} + (1-\alpha) \theta, \qquad
\alpha = 1 - (0.2 \times N_\text{epochs})^{-1}.
\end{align*}

During validation and testing, we use the EMA weights instead of the regular model weights.

\subsubsection{Optimizer}
\label{sm:optimizer}
In all experiments, we use the AdamW optimizer \cite{loshchilov2017}. The learning rate is scaled with the square root of the batch size \cite{you2019,malladi2022,li2024d,shuai2024}; that is, we set the learning rate equal to the base learning rate multiplied by the square root of the batch size. However, for a given dataset and training phase (i.e., SSL pre-training, probing, or fine-tuning), we keep the batch size and learning rate fixed across experiments to avoid any confounding factors.

Across all phases—SSL pre-training, probing, and fine-tuning—we use a cosine decay learning rate scheduler with a single cycle and a warm-up period. This approach is common in recent SSL studies with Transformers in computer vision \cite{caron2021,he2021,xie2021,oquab2023}. We implement this using \verb|torch.optim.lr_scheduler.OneCycleLR| with its default annealing strategy and a warm-up for 20\% of the total epochs.

During SSL pre-training and probing, the learning rate is annealed down to $1 \times 10^{-4}$ times its maximum value. During fine-tuning, it is instead annealed only to half of the maximum learning rate. This helps fully leverage the EMA strategy as weight averaging (i) naturally reduces noise \cite{morales2024} and (ii) benefits from greater model diversity along the training trajectory when the learning rate remains sufficiently high.

\subsection{Experimental Details Specific to MAEs/MAESTRO}
\label{sm:MAE_MAESTRO}

In this subsection, we provide experimental details specific to MAEs/MAESTRO/ViTs. We detail the masking strategy (\cref{sm:masking}) and the selection of band groups for patch-group-wise normalization (\cref{sm:band_groups}). We also provide details on minor adaptations of MAESTRO to enable cross-dataset transfer (\cref{sm:cross_dataset}).

\subsubsection{Masking}
\label{sm:masking}

\paragraph{Masking Strategy.} We adopt a masking strategy that proceeds in two stages:
\begin{enumerate}[label=(\roman*)]
\item \emph{Structured Masking:}
\begin{enumerate}[label=(\alph*)]
\item \emph{Modality structure:} Mask each modality with a fixed probability of $0.25$;
\item  \emph{Spatial structure:} Within each modality, mask each spatial position with a fixed probability of $0.25$;
\item  \emph{Temporal structure:} Within each modality, mask each temporal position with a fixed probability of $0.25$.
\end{enumerate}
\item \emph{Unstructured Masking:}
Adjust the masking from step (i) to match an overall 75\% masking ratio:
\begin{itemize}
\item If step (i) results in too few masked tokens, randomly mask additional unmasked tokens;
\item If step (i) results in too many masked tokens, randomly unmask some of them.
\end{itemize}
\end{enumerate}

\paragraph{Comparison with masking strategies from previous works.} Our masking strategy has similarities with previous works.

Step (i-a), which introduces modality-structured masking, is conceptually similar to the band-masking strategy in Fomo-Net~\cite{bountos2025}, although a direct equivalence would require treating each spectral band as a distinct modality.

Step (i-b), which introduces spatially-structured masking, aligns with the consistent masking in SatMAE~\cite{cong2022} and the tube masking in VideoMAE~\cite{tong2022}. It also resembles the approach in SeaMo~\cite{li2024b}, but with a key distinction: SeaMo enforces consistency across modalities at a single time step, whereas we enforce consistency across time steps within a single modality.

Step (i-c), which introduces temporally-structured masking, corresponds to the \say{Timesteps} strategy in Presto~\cite{tseng2023}.

Our combined masking strategy shares similarities with those in AnySat~\cite{astruc2024b}, EarthMAE~\cite{velazquez2025}, and Galileo~\cite{tseng2025}, but still differs in important ways. AnySat and Galileo apply only spatially- and temporally-structured masking, while EarthMAE uses only spatially-structured masking. Moreover, the way in which we integrate the structured and unstructured stages is distinct from these methods.

\subsubsection{Band Groups}
\label{sm:band_groups}

Here, we provide details on our selection of spectral band groups for patch-group-wise normalization.

\begin{figure}[h]
  \centering

  \begin{subfigure}[b]{0.32\textwidth}
    \centering
    \includegraphics[width=\textwidth]{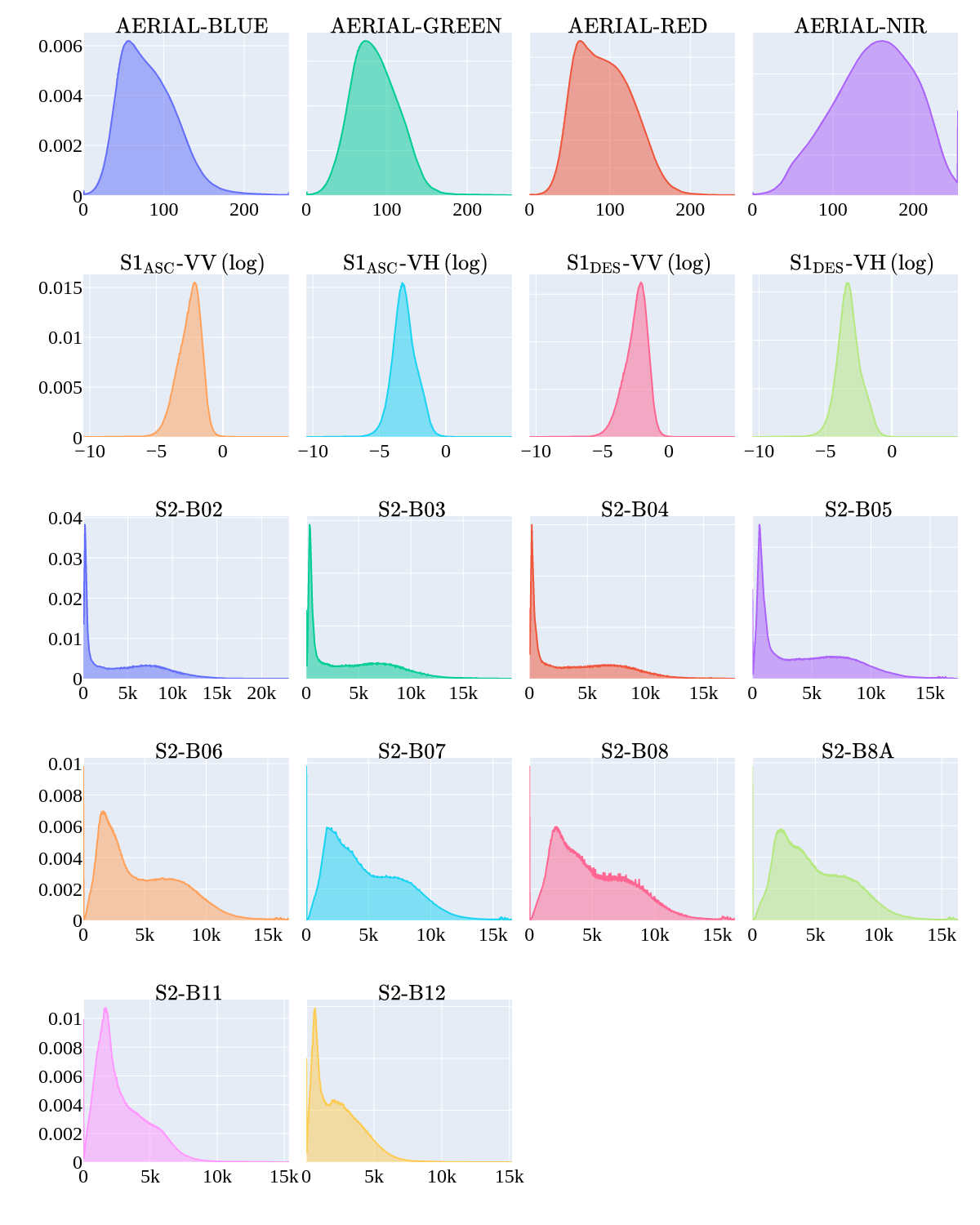}
    \caption{TreeSatAI-TS}
    \label{fig:histos_treesat}
  \end{subfigure}
  \hspace{0.01\textwidth}
  \begin{subfigure}[b]{0.32\textwidth}
    \centering
    \includegraphics[width=\textwidth]{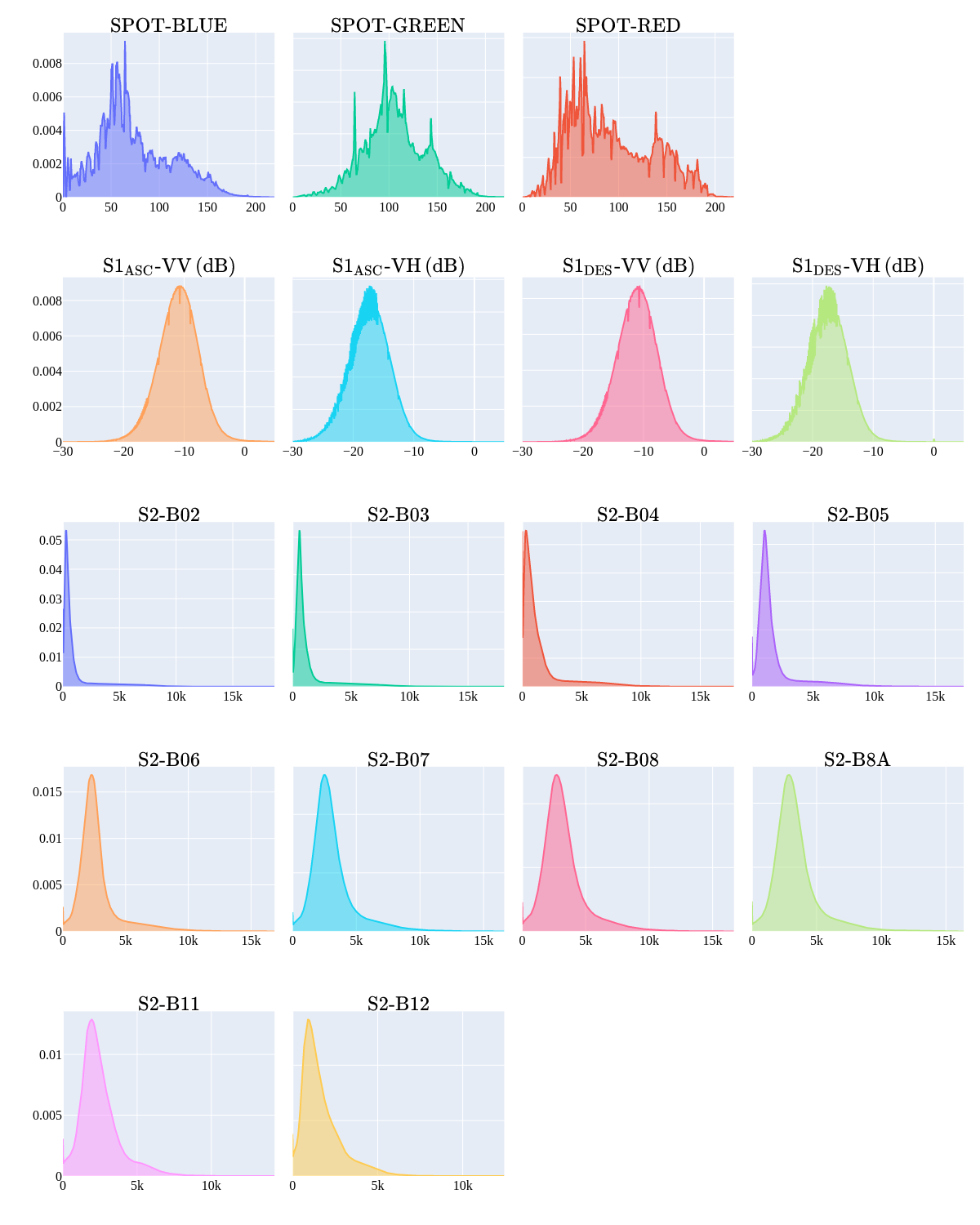}
    \caption{PASTIS-HD}
    \label{fig:histos_pastis}
  \end{subfigure}
  \hspace{0.01\textwidth}
  \begin{subfigure}[b]{0.32\textwidth}
    \centering
    \includegraphics[width=\textwidth]{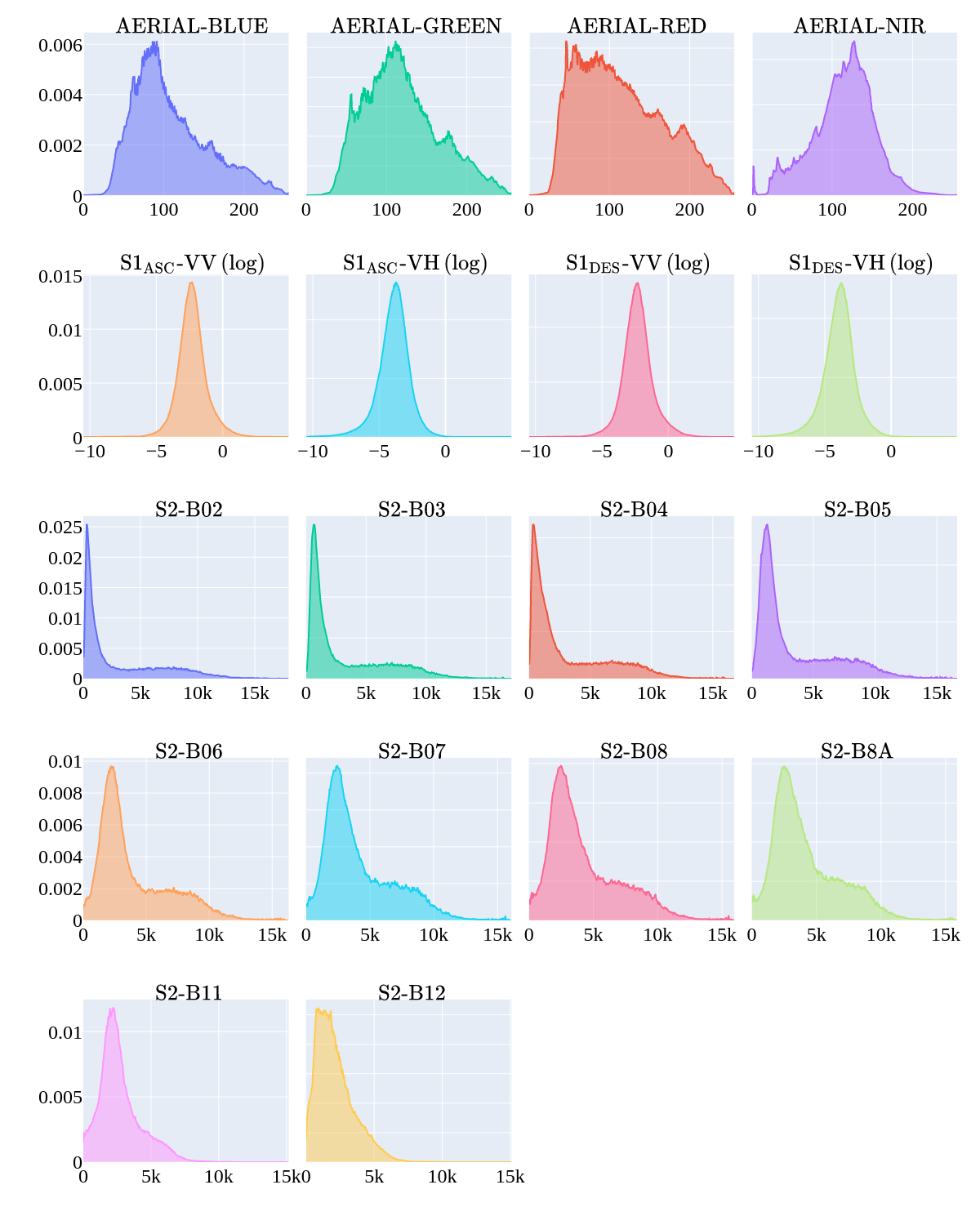}
    \caption{FLAIR-HUB}
    \label{fig:histos_flair}
  \end{subfigure}

  \caption{\centering \textbf{Per-band histograms for VHR, Sentinel-1, and Sentinel-2 modalities on the TreeSatAI-TS, PASTIS-HD, and FLAIR-HUB datasets.}}
  \label{fig:band_histograms}
\end{figure}

We start by computing per-band histograms for the VHR, Sentinel-1, and Sentinel-2 modalities on TreeSatAI-TS, PASTIS-HD, and FLAIR-HUB. To reduce computational cost, a toy version of FLAIR-HUB containing 250 tiles was used for histogram computation. The resulting histograms are shown in \cref{fig:histos_treesat}, ~\cref{fig:histos_pastis}, and~\cref{fig:histos_flair}.

For VHR aerial imagery, the RED, GREEN, and BLUE bands exhibit relatively similar histograms, whereas the NIR band shows a distinct distribution.

For Sentinel-1, the histograms of VV and VH differ markedly, in line with their differing polarization responses; VH generally exhibits lower backscatter than VV.

For Sentinel-2, the ten bands cluster into three natural groups with strong intra-group similarity but higher inter-group variation.

\begin{table}[h]
\centering
\footnotesize
\caption{\textbf{Spectral band groups selected for patch-group-wise normalization.}}
\label{tab:band_groups}
\begin{tabular}{l l l l}
\toprule
Modality & Group 1 & Group 2 & Group 3 \\
\midrule
Aerial & RED, GREEN, BLUE & NIR & \\
SPOT   & RED, GREEN, BLUE & & \\
S1\textsubscript{ASC} / S1\textsubscript{DES} & VV & VH & \\
S2     & B02, B03, B04, B05 & B06, B07, B08, B8A & B11, B12 \\
\bottomrule
\end{tabular}
\end{table}

Based on these insights, we define the spectral band groups detailed in \cref{tab:band_groups} for patch-group-wise normalization.

Note that the Sentinel-2 band groups follow the natural wavelength order, which does not precisely align with the order of spatial resolutions (e.g., band B05 has 20 m resolution, while B08 has 10 m). This contrasts with some prior approaches to S2 band grouping \cite{cong2022, irvin2023}.

\subsubsection{Cross-Dataset Transfer}
\label{sm:cross_dataset}

Here, we detail the adaptations made for cross-dataset transfer compared to the intra-dataset setting.

First, we ensure that the patch sizes $P_m$ for each modality match between the pre-training and fine-tuning datasets in cross-dataset MAESTRO. This allows the patchification layers to be fully transferred across datasets. During fine-tuning, we adjust the image size $I_m$ for each modality to maintain the same token budget as in intra-dataset MAESTRO, ViTs, and adapted baseline FMs while using the fixed patch size $P_m$, ensuring a fair comparison.

Second, \emph{fixed cross-dataset grids} for positional encodings are defined based on the ground sampling distance, following the approach in Scale-MAE \cite{reed2023}.

Finally, when pre-training on S2-NAIP urban, we generate surrogate modalities for aerial and SPOT imagery via resampling of NAIP imagery (see \cref{tab:s2naip_hyperparams}). These surrogate modalities are transferred to the corresponding modalities in the downstream tasks. Additionally, we do not distinguish the Sentinel-1 modality by orbit during pre-training, as this information is unavailable. During fine-tuning on TreeSatAI-TS, PASTIS-HD, and FLAIR-HUB, we retain the separate Sentinel-1 modalities by orbit, using a single patchification layer initialized from the pre-trained Sentinel-1 layer. The Sentinel-1 encoder is subsequently transferred to the grouped encoder for Sentinel-1 orbits, consistent with MAESTRO’s \emph{inter-group} fusion mode, where these modalities are grouped.

\subsection{Experimental Details Specific to Adapted Baseline FMs}
\label{sm:FMs_details}

In this subsection, we describe the baseline FMs that we use as baselines in our comparative evaluation of MAESTRO, along with their extensive adaptations. 

The selected models include DINO-v2~\cite{oquab2023}, which is pre-trained on natural images, as well as DINO-v2 sat.~\cite{tolan2024}, CROMA~\cite{fuller2023}, DOFA~\cite{xiong2024}, SatMAE~\cite{cong2022}, and Prithvi-EO 2.0~\cite{szwarcman2024}, which are pre-trained on EO data. 

\subsubsection{Limitations in the Original Models}

In their original configurations, the selected baseline FMs exhibit several limitations:
\begin{itemize}
\item \textbf{CROMA/DOFA.} CROMA natively supports multiple modalities, but it is monotemporal. DOFA also supports different modalities, but only through parameter sharing and not through multimodal fusion. Additionally, DOFA is monotemporal.

\item \textbf{DINO-v2/DINO-v2 sat.} DINO-v2 and DINO-v2 sat are strictly monomodal and monotemporal. Additionally, DINO-v2 and DINO-v2 sat were only pre-trained on RGB images.

\item \textbf{Prithvi-EO 2.0/SatMAE.} Prithvi-EO 2.0 is natively multitemporal, but its pre-training is limited to four Sentinel-2 dates. SatMAE also supports multitemporality, but its reference implementation and its pre-training are limited to three Sentinel-2 dates. Both models were pre-trained on a limited subset of Sentinel-2 bands: six bands (RGB, NIR, SWIR1, SWIR2) for Prithvi-EO 2.0, and three bands (RGB) for SatMAE.
\end{itemize}

We refer to \cref{tab:comparison_high_level} for more details on some of these limitations.

\subsubsection{Overview of Adaptations}

To optimize the performance of these baseline FMs, we introduce several key adaptations:

\begin{itemize}
    \item \textbf{Multimodal tokenization:} 
    \begin{itemize}
	  \item Multiple modalities are processed via parallel, modality-specific tokenizers. 
  	\end{itemize}
    \item \textbf{Additional bands incorporation:} 
    \begin{itemize}
	  \item Tokenizers are modified to support bands not seen during pre-training, while retaining weights for previously available bands to minimize transfer disruption.
  	\end{itemize}
    \item \textbf{Mulimodal/multitemporal fusion:} 
    \begin{itemize}
    \item For DINO-v2, DINO-v2 sat, DOFA, we add late fusion across modalities and time steps, using the \textit{shared} and \textit{monotemp} modes (\cref{sec:architecture}).
    \item For CROMA, we adopt specialized multimodal and multitemporal fusion modes.
    \item For SatMAE, Prithvi-EO 2.0, we apply early multitemporal fusion as in the original models, but we allow the handling of arbitrary number of Sentinel-2 dates.
	\end{itemize}
    \item \textbf{Temporal encoding:} 
     \begin{itemize}
	\item For DINO-v2, DINO-v2 sat, DOFA, CROMA, we inject the temporal encodings from \cref{sm:encodings} into the embedded tokens \emph{after} the encoder. This enables to limit transfer disruption.
	\item For SatMAE, Prithvi-EO 2.0, we use the temporal encodings from the original models.
	\end{itemize}
    \item \textbf{Input resizing:} 
    \begin{itemize}
      \item For all baseline FMs, input images are resized to match MAESTRO’s token grid while preserving the original patch size (see \cref{tab:treesat_hyperparams,tab:pastis_hyperparams,tab:flair_hyperparams}).
  	\end{itemize}
\end{itemize}

All adapted baseline FMs natively use ViT backbones~\cite{dosovitskiy2020}, as in MAESTRO. We also use the same classification and segmentation heads as in MAESTRO.

In \cref{tab:baseline_details}, we report the model sizes as well as the pre-training modalities and the modalities selected for fine-tuning for each adapted baseline FM.

\begin{table*}[ht!]
\centering
\footnotesize
\caption{\textbf{Overview of the adapted baseline FMs.} Model sizes, pre-training modalities, and modalities selected for fine-tuning.}
\label{tab:baseline_details}
\begin{tabular}{ll lcccccc}
\toprule
\multirow{2.6}{*}{Model} & \multirow{2.6}{*}{\makecell[c]{Model\\size}} & \multirow{2.6}{*}{Pre-training modalities} & \multicolumn{4}{c}{Fine-tuning modalities} \\
 \cmidrule(lr){4-7}
& & & VHR & $\text{S1}_\text{ASC}$/$\text{S1}_\text{DES}$ & S2 & DEM/DSM \\
\midrule
                   DINO-v2 \cite{oquab2023}           & Base  & Natural RGB images (LVD-142M dataset) & \cmark & \xmark & \cmark & \xmark \\
\rowcolor{black!5} DINO-v2 sat. \cite{tolan2024} & Large & Maxar Vivid2 mosaic imagery & \cmark & \xmark & \cmark & \xmark \\
                   DOFA \cite{xiong2024}              & Base  & Sentinel-1, Sentinel-2, Gaofen, NAIP, EnMaP & \cmark & \cmark & \cmark & \xmark \\
\rowcolor{black!5} CROMA \cite{fuller2023}            & Base  & Sentinel-1, Sentinel-2 (SSL4EO dataset) & \xmark & \cmark & \cmark & \xmark \\
                   Prithvi-EO-2.0 \cite{szwarcman2024} & Large & Sentinel-2, Landsat & \xmark & \xmark & \cmark & \xmark \\
\rowcolor{black!5} SatMAE \cite{cong2022}             & Large & Sentinel-2 (fMoW RGB+Sentinel dataset) & \xmark & \xmark & \cmark & \xmark \\
\bottomrule
\end{tabular}
\end{table*}

\subsubsection{DINO-v2}
\label{sm:adapt_DinoV2}
DINO-v2~\cite{oquab2023} is a FM pre-trained on natural RGB images using a SSL teacher–student distillation framework, which combines global (image-level) and local (patch-level) objectives. Due to the domain gap between natural and EO imagery, transferring DINO-v2 to EO tasks is non-trivial.

To adapt it to our multimodal setting, we modify its patchification operation to support both VHR and Sentinel-2 inputs using two parallel patchification layers:
\begin{itemize}
\item On PASTIS-HD, FLAIR\#2, and FLAIR-HUB, we retain the pre-trained RGB channel weights, while additional channels are initialized with near-zero values to minimize disruption, following~\cite{fogel2025}. 
\item On TreeSatAI-TS, however, we found it beneficial to map the pre-trained RGB weights to the infrared colors (IRC) channels NIR, RED, and GREEN in the downstream task, while initializing the BLUE channel weights with near-zero values. We attribute this beneficial effect to two factors: (i) representations learned on natural RGB images transfer well to IRC aerial imagery, and (ii) the NIR channel plays a decisive role in TreeSatAI-TS, benefiting from being mapped to the transferred weights.
\end{itemize}
We experiment with two fusion modes for handling multimodality and multitemporality:
\begin{enumerate}[label=(\roman*)]
\item \textit{shared}, where a single encoder (initialized from DINO-v2’s weights) processes all modalities and time steps independently, with shared weights across them,
\item \textit{monotemp}, where modality-specific encoders (also initialized from DINO-v2) process each modality and time step independently.
\end{enumerate}

Since DINO-v2 lacks native temporal handling, the temporal encodings from \cref{sm:encodings} are injected into the post-encoder token embeddings. The class token is discarded.

\subsubsection{DINO-v2 sat}

DINO-v2 sat~\cite{tolan2024} is based on the same architecture as DINO-v2 but pre-trained on Maxar’s very high-resolution RGB satellite imagery. We apply the same adaptation protocol for this model as for DINO-v2.

\subsubsection{DOFA}
DOFA~\cite{xiong2024} is a FM pre-trained on multi-source EO data using the MAE framework and a dynamic channel handling via a hypernetwork that generates patch embedding weights based on input wavelengths (see Table~\ref{tab:dofa_wavelengths}). While DOFA is inherently flexible, adaptations are required to support multimodal and multitemporal fusion.

We implement four parallel patchification layers (initialized from the original patchification layer) to handle VHR, Sentinel-2, and Sentinel-1 in both ascending and descending orbits. 

As with DINO-v2 and DINO-v2 sat, we explore both \textit{shared} and \textit{monotemp} fusion modes for handling multimodality and multitemporality. 

Since the original model lacks native temporal handling, the temporal encodings from \cref{sm:encodings} are again injected into the post-encoder token embeddings.

\begin{table}[ht]
\centering
\footnotesize
\caption{\textbf{DOFA's wavelengths per modality.}}
\label{tab:dofa_wavelengths}
\begin{tabular}{l l l}
\toprule
Modality & Bands & Wavelengths \\
\midrule
Aerial        & RED, GREEN, BLUE, NIR & 0.64, 0.56, 0.48, 0.81 \\
SPOT          & RED, GREEN, BLUE      & 0.66, 0.56, 0.48 \\
S1\textsubscript{ASC} & VV, VH         & 5.405, 5.405 \\
S1\textsubscript{DES} & VV, VH         & 5.405, 5.405 \\
S2            & B02, B03, B04, B05, B06, B07, B08, B8A, B11, B12 & 0.490, 0.560, 0.665, 0.705, 0.740, 0.783, 0.842, 0.865, 1.610, 2.190 \\
\bottomrule
\end{tabular}
\end{table}

\subsubsection{CROMA}
\label{sm:adapt_CROMA}

CROMA \cite{fuller2023} is a FM specifically designed for Sentinel imagery, featuring separate encoders for Sentinel-1 and Sentinel-2, coupled via a cross-encoder that fuses their intermediate representations. CROMA is pre-trained on SSL4EO dataset. Since we fine-tune CROMA on its original modalities, only minimal adaptations are required.

We consider two fusion modes for handling multimodality and multitemporality: 
\begin{enumerate}[label=(\roman*)]
\item \textit{late-croma} (cross-encoder disabled), where monotemporal Sentinel-1 and Sentinel-2 encoder outputs are concatenated across modalities and time steps before the classification/segmentation heads;
\item \textit{inter-croma} (default), where monotemporal Sentinel-1 and Sentinel-2 encoder outputs are grouped into pairs, fused via the cross-encoder, and finally concatenated across pairs before the classification/segmentation heads.
\end{enumerate}

In each case, the time series for Sentinel-1 in ascending and descending orbits are concatenated and passed as a single sequence to the Sentinel-1 encoder.

As with other baseline FMs, the temporal encodings from \cref{sm:encodings} are injected post-encoder.

Note that CROMA remains intrinsically monotemporal: \textit{late-croma} performs only late multitemporal fusion, while \textit{inter-croma} performs intermediate multitemporal fusion within individual Sentinel-1/Sentinel-2 pairs, but late fusion across pairs.

\subsubsection{Prithvi-EO}
\label{sm:adapt_prithvi}
Prithvi-EO 2.0 \cite{szwarcman2024} is a geospatial FM pre-trained on a corpus of NASA's Harmonized Landsat Sentinel-2 (HLS) data at 30-meter resolution. The architecture is an adaptation of the MAE to satellite time series, employing token-based multitemporal fusion. The standard 2D patch and positional embeddings of the MAE architecture are replaced with 3D counterparts to natively handle spatiotemporal inputs. 

In its original form, the model processes only six Sentinel-2 bands: Red, Green, Blue, NIR, SWIR 1 \& 2. To extend it to additional bands, we adapt the patchification layer to accept new bands. The pre-trained channel weights are retained, while the extra channels are initialized with near-zero values to minimize transfer disruption, following~\cite{fogel2025}.

Although the original model was pre-trained on sequences of four Sentinel-2 dates, the official TerraTorch implementation does not impose a limitation on the number of dates. We leverage this to enable the model to process an arbitrary number of time steps.

We retain the same temporal encodings as in the original model.

\subsubsection{SatMAE}
Temporal SatMAE \cite{cong2022} is a pre-training framework that adapts the MAE to satellite time series, employing token-based multitemporal fusion.

In its original form, the model processes only three Sentinel-2 bands (RGB). To extend it to additional bands, we adapt the patchification layer to accept new bands. The pre-trained RGB channel weights are retained, while the extra channels are initialized with near-zero values to minimize transfer disruption, following~\cite{fogel2025}.

A further limitation of the original model is its restriction to three Sentinel-2 dates. We remove this constraint and enable the model to process an arbitrary number of time steps.

We retain the same temporal encodings as in the original model.

\subsection{Hyperparameter Tables}
\label{sm:hyperparam_tables}

In this subsection, we provide tables reporting the exhaustive list of our hyperparameter values.

\begin{table}[H]
\centering
\footnotesize
\caption{\textbf{TreeSatAI-TS's hyperparameters.}}
\begin{tabular}{lllll}
\toprule
& \multicolumn{4}{l}{\makecell[c]{\textit{Spatial extent of original tiles: 60~m} \\ \textit{Spatial extent of crops: 60~m}}} \\
\midrule
 & Aerial & $\text{S1}_\text{ASC}$ & $\text{S1}_\text{DES}$ & S2 \\
\midrule
Dataset's original resolution (m) & 0.2 & 10 & 10 & 10 \\
Image size ($I_m$) & & & & \\
\quad MAESTRO-intra/MAE/ViT & 300 & 6 & 6 & 6 \\
\quad MAESTRO-cross & 240 & 6 & 6 & 6 \\
\quad DINO-v2 & 210 & \xmark & \xmark & 42 \\
\quad DINO-v2 sat. & 240 & \xmark & \xmark & 48 \\
\quad DOFA & 240 & 48 & 48 & 48 \\
\quad CROMA & \xmark & 24 & 24 & 24 \\
\quad Prithvi-EO-2.0/SatMAE & \xmark & \xmark & \xmark & 48 \\
Patch size ($P_m$) & & & & \\
\quad MAESTRO-intra/MAE/ViT & 20 & 2 & 2 & 2 \\
\quad MAESTRO-cross & 16 & 2 & 2 & 2 \\
\quad DINO-v2 & 14 & \xmark & \xmark & 14 \\
\quad DINO-v2 sat. & 16 & \xmark & \xmark & 16 \\
\quad DOFA & 16 & 16 & 16 & 16 \\
\quad CROMA & \xmark & 8 & 8 & 8 \\
\quad Prithvi-EO-2.0/SatMAE & \xmark & \xmark & \xmark & 16 \\
Number of temporal bins ($D_m$) & 1 & 4 & 4 & 16 \\
Number of channels ($C_m$) & 4 & 2 & 2 & 10 \\
Multiplicative normalization factor & 255 & 5 & 5 & $5 \times 10^3$ \\
Use cloud/snow mask &  & & & \cmark (mask proba. $>0$) \\
\midrule
\multirow{3}{*}{\makecell[tl]{MAE/MAESTRO's band groups ($\mathcal{G}_m$)}}
  & RED, GREEN, BLUE & VV & VV & B02, B03, B04, B05 \\
  & NIR              & VH & VH & B06, B07, B08, B8A \\
  &                  &    &    & B11, B12 \\
\bottomrule
\end{tabular}
\label{tab:treesat_hyperparams}
\end{table}

\begin{table}[H]
\centering
\footnotesize
\caption{\textbf{PASTIS-HD's hyperparameters.}}
\begin{tabular}{lllll}
\toprule
& \multicolumn{4}{l}{\makecell[c]{\textit{Spatial extent of original tiles: 1280~m} \\ \textit{Spatial extent of crops: 160~m} \\ \textit{Spatial resolution of token grid of reference: 20~m} }} \\
\midrule
 & SPOT & $\text{S1}_\text{ASC}$ & $\text{S1}_\text{DES}$ & S2 \\
\midrule
Dataset's original resolution (m) & 1 & 10 & 10 & 10 \\
Image size ($I_m$) & & & & \\
\quad MAESTRO-intra/MAE/ViT & 160 & 16 & 16 & 16 \\
\quad MAESTRO-cross & 160 & 16 & 16 & 16 \\
\quad DINO-v2 & 140 & \xmark & \xmark & 112 \\
\quad DINO-v2 sat. & 160 & \xmark & \xmark & 128 \\
\quad DOFA & 160 & 128 & 128 & 128 \\
\quad CROMA & \xmark & 64 & 64 & 64 \\
\quad Prithvi-EO-2.0/SatMAE & \xmark & \xmark & \xmark & 128 \\
Patch size ($P_m$) & & & & \\
\quad MAESTRO-intra/MAE/ViT & 16 & 2 & 2 & 2 \\
\quad MAESTRO-cross & 16 & 2 & 2 & 2 \\
\quad DINO-v2 & 14 & \xmark & \xmark & 14 \\
\quad DINO-v2 sat. & 16 & \xmark & \xmark & 16 \\
\quad DOFA & 16 & 16 & 16 & 16 \\
\quad CROMA & \xmark & 8 & 8 & 8 \\
\quad Prithvi-EO-2.0/SatMAE & \xmark & \xmark & \xmark & 16 \\
Number of temporal bins ($D_m$) & 1 & 4 & 4 & 16 \\
Number of channels ($C_m$) & 3 & 2 & 2 & 10 \\
Multiplicative normalization factor & 255 & 20 & 20 & $1 \times 10^4$ \\
Use cloud/snow mask &  & & & \xmark \;(not avail.) \\
\midrule
\multirow{3}{*}{\makecell[tl]{MAE/MAESTRO's band groups ($\mathcal{G}_m$)}}
  & RED, GREEN, BLUE & VV & VV & B02, B03, B04, B05 \\
  &                  & VH & VH & B06, B07, B08, B8A \\
  &                  &    &    & B11, B12 \\
\bottomrule
\end{tabular}
\label{tab:pastis_hyperparams}
\end{table}

\begin{table}[H]
\centering
\footnotesize
\caption{\textbf{FLAIR\#2's and FLAIR-HUB's hyperparameters.}}
\begin{tabular}{llllll}
\toprule
& \multicolumn{5}{l}{\makecell[c]{\textit{Spatial extent of original tiles: 102.4~m} \\ \textit{Spatial extent of crops: 102.4~m} \\ \textit{Spatial resolution of token grid of reference: 3.2~m}}} \\
\midrule
 & Aerial & DEM/DSM & $\text{S1}_\text{ASC}$ & $\text{S1}_\text{DES}$ & S2 \\
\midrule
Present in FLAIR\#2 & \cmark & \cmark & \xmark & \xmark & \cmark \\
\midrule
Dataset's original resolution (m) & 0.2 & 0.2 & 10.24 & 10.24 & 10.24 \\
Image size ($I_m$) & & & & & \\
\quad MAESTRO-intra/MAE/ViT & 512 & 512 & 10 & 10 & 10 \\
\quad MAESTRO-cross & 512 & \xmark & 10 & 10 & 10 \\
\quad DINO-v2 & 448 & \xmark & \xmark & \xmark & 70 \\
\quad DINO-v2 sat. & 512 & \xmark & \xmark & \xmark & 80 \\
\quad DOFA & 512 & \xmark & 80 & 80 & 80 \\
\quad CROMA & \xmark & \xmark & 40 & 40 & 40 \\
\quad Prithvi-EO-2.0/SatMAE & \xmark & \xmark & \xmark & \xmark & 80 \\
Patch size ($P_m$) & & & & & \\
\quad MAESTRO-intra/MAE/ViT & 16 & 32 & 2 & 2 & 2 \\
\quad MAESTRO-cross & 16 & \xmark & 2 & 2 & 2 \\
\quad DINO-v2 & 14 & \xmark & \xmark & \xmark & 14 \\
\quad DINO-v2 sat. & 16 & \xmark & \xmark & \xmark & 16 \\
\quad DOFA & 16 & \xmark & 16 & 16 & 16 \\
\quad CROMA & \xmark & \xmark & 8 & 8 & 8 \\
\quad Prithvi-EO-2.0/SatMAE& \xmark & \xmark & \xmark & \xmark & 16 \\
Number of temporal bins ($D_m$) & 1 & 1 & 4 & 4 & 16 \\
Number of channels ($C_m$) & 4 & 2 & 2 & 2 & 10 \\
Multiplicative normalization factor & 255 & $1 \times 10^3$ & 5 & 5 & $5 \times 10^3$ \\
Use cloud/snow mask &  &  & & & \cmark (mask proba. $>0$) \\
\midrule
\multirow{3}{*}{\makecell[tl]{MAE/MAESTRO's band groups ($\mathcal{G}_m$)}}
  & RED, GREEN, BLUE & DEM, DSM & VV & VV & B02, B03, B04, B05 \\
  & NIR              &          & VH & VH & B06, B07, B08, B8A \\
  &                  &          &    &    & B11, B12 \\
\bottomrule
\end{tabular}
\label{tab:flair_hyperparams}
\end{table}

\begin{table}[H]
\centering
\footnotesize
\caption{\textbf{S2-NAIP urban's hyperparameters.}}
\begin{tabular}{lllll}
\toprule
& \multicolumn{4}{l}{\makecell[c]{\textit{Spatial extent of original tiles: 640~m} \\ \textit{Spatial extent of crops: 120~m}}} \\
\midrule
 & NAIP -- SPOT surrogate & NAIP -- Aerial surrogate & S1 & S2  \\
 \midrule
Dataset's original resolution (m) & 1.25 & 1.25 & 10 & 10 \\
Image size ($I_m$) & & & & \\
\quad MAESTRO-cross & 384 & 128 & 12 & 12 \\
Patch size ($P_m$) & & & & \\
\quad MAESTRO-cross & 16 & 16 & 2 & 2 \\
Number of temporal bins ($D_m$) & 1 & 1 & 4 & 16 \\
Number of channels ($C_m$) & 4 & 3 & 2 & 10 \\
Multiplicative normalization factor & 255 & 255 & 20 &  $5 \times 10^3$ \\
Use cloud/snow mask &  &  &  & \xmark \;(not avail.) \\
\midrule
\multirow{3}{*}{\makecell[tl]{MAE/MAESTRO's band groups ($\mathcal{G}_m$)}} 
  & RED, GREEN, BLUE & RED, GREEN, BLUE & VV & B02, B03, B04, B05 \\
  & NIR               &                  & VH & B06, B07, B08, B8A \\
  &                   &                  &    & B11, B12 \\
\bottomrule
\end{tabular}
\label{tab:s2naip_hyperparams}
\end{table}

\begin{table}[H]
\centering
\footnotesize
\caption{\textbf{Data augmentation/regularization hyperparameters.}}
\begin{tabular}{lll}
\toprule
Augmentation/Regularization & Phase used & Hyperparameter \\
\midrule
Random spatial cropping & pre-train/probe/fine-tune \\ 
Random time step selection & pre-train/probe/fine-tune \\ 
D4 augmentation & pre-train/probe/fine-tune & \\
\midrule
EMA \cite{morales2024} & Fine-tune & $\alpha = 1 - (0.2 \times N_\text{epochs})^{-1}$ \\
\bottomrule
\end{tabular}
\label{tab:augreg_hyperparams}
\end{table}

\begin{table}[H]
\centering
\footnotesize
\caption{\textbf{Optimizer hyperparameters.}}
\begin{tabular}{lllllll}
\toprule
Phase & Hyperparameter & TreeSatAI-TS & PASTIS-HD & FLAIR\#2 & FLAIR-HUB & S2-NAIP urban \\
\midrule
 & Optimizer & AdamW & AdamW & AdamW & AdamW & AdamW \\
 & LR schedule & cosine decay & cosine decay & cosine decay & cosine decay & cosine decay \\
 & Warmup fraction & 20\% & 20\% & 20\% & 20\% & 20\% \\
 & Weight decay & $1 \times 10^{-2}$ & $1 \times 10^{-2}$ & $1 \times 10^{-2}$ & $1 \times 10^{-2}$ & $1 \times 10^{-2}$ \\
 & $\beta_1$ & 0.9 & 0.9 & 0.9 & 0.9 & 0.9 \\
 & $\beta_2$ & 0.99 & 0.99 & 0.99 & 0.99 & 0.99 \\
\midrule
\multirow{10}{*}{Pre-train} 
 & Base learning rate ($\times \sqrt{3}$) & $3 \times 10^{-5}$ & $3 \times 10^{-5}$ & $3 \times 10^{-5}$ & $3 \times 10^{-5}$ & $1 \times 10^{-5}$ \\
 & Final div. factor   & $1 \times 10^{4}$ & $1 \times 10^{4}$ & $1 \times 10^{4}$ & $1 \times 10^{4}$ & $1 \times 10^{4}$ \\
 & Batch size & & & & & \\
  & \quad Dataset frac. 100\% & 96 & 72 & 72 & 144 & 512 \\
  & \quad Dataset frac. 20\%  & 96 & 72 & -- & 72 & -- \\
  & \quad Dataset frac. 5\%   & 96 & 72 & -- & 72 & -- \\
 & Epochs & & & & & \\
 & \quad Dataset frac. 100\% & 100 & 100 & 100 & 100 & 15 \\
 & \quad Dataset frac. 20\%  & 200 & 200 & --  & 100 & -- \\
 & \quad Dataset frac. 5\%   & 400 & 400 & --  & 200 & -- \\
\midrule
\multirow{10}{*}{Probe} 
 & Base learning rate {\scriptsize($\times \sqrt{3}$)} & $1 \times 10^{-5}$ & $1 \times 10^{-5}$ & $1 \times 10^{-5}$ & $1 \times 10^{-5}$ & -- \\
 & Final div. factor   & $1 \times 10^{4}$ & $1 \times 10^{4}$ & $1 \times 10^{4}$ & -- \\
 & Batch size & & & & & \\
  & \quad Dataset frac. 100\% & 96 & 48 & 48 & 96 & -- \\
  & \quad Dataset frac. 20\%  & 96 & 48 & -- & 48 & -- \\
  & \quad Dataset frac. 5\%   & 96 & 48 & -- & 48 & -- \\
 & Epochs & & & & & \\
 & \quad Dataset frac. 100\% & 10 & 10 & 15 & 15 & -- \\
 & \quad Dataset frac. 20\%  & 20 & 20 & -- & 15 & -- \\
 & \quad Dataset frac. 5\%   & 40 & 40 & -- & 30 & -- \\
\midrule
\multirow{10}{*}{Fine-tune} 
 & Base learning rate ($\times \sqrt{3}$) & $1 \times 10^{-5}$ & $1 \times 10^{-5}$ & $1 \times 10^{-5}$ & $1 \times 10^{-5}$ & -- \\
 & Final div. factor   & 2 & 2 & 2 & 2 & -- \\
 & Batch size & & & & & \\
  & \quad Dataset frac. 100\% & 96 & 48 & 48 & 96 & -- \\
  & \quad Dataset frac. 20\%  & 96 & 48 & -- & 48 & -- \\
  & \quad Dataset frac. 5\%   & 96 & 48 & -- & 48 & -- \\
 & Epochs & & & & & \\
 & \quad Dataset frac. 100\% & 50  & 50  & 100 & 100 & -- \\
 & \quad Dataset frac. 20\%  & 100 & 100 & --  & 100 & -- \\
 & \quad Dataset frac. 5\%   & 200 & 200 & --  & 200 & -- \\
\bottomrule
\end{tabular}
\label{tab:opt_all}
\end{table}

\begin{table}[H]
\centering
\footnotesize
\caption{\textbf{SSL hyperparameters.}}
\begin{tabular}{ll}
\toprule
Hyperparameter &  \\
\midrule
\makecell[l]{Reconstruction loss\\ \,} & \makecell[l]{%
$L_1$ w/ patch-group-wise normalization \\
(for some ablations: $L_1$ w/o normalization or $L_1$ w/ patch-wise normalization)
} \\
\midrule
Masking ratio & 75\% \\
\makecell[l]{Probability modality-structured masking} & \makecell[l]{0.25 (if not disabled)}  \\
\makecell[l]{Probability temporally-structured masking} & \makecell[l]{0.25 (if not disabled)}   \\
\makecell[l]{Probability spatially-structured masking} & \makecell[l]{0.25 (if not disabled)}  \\
\bottomrule
\end{tabular}
\label{tab:ssl_hyperparams}
\end{table}

%% file: 7_detailed_results.tex
\section{Detailed Results}
\label{sm:detailed_results}

In this section, we report the detailed results on multimodal and multitemporal fusion (\cref{sm:fusion_modes}), multispectral fusion and target normalization (\cref{sm:multispectral_comparison}), and scaling with respect to pre-training and fine-tuning dataset size (\cref{sm:Scaling_byDataset_Size}). 

\subsection{Multimodal/Multitemporal Fusion}
\label{sm:fusion_modes}

The detailed numbers of the evaluation of the different multimodal and multitemporal fusion modes with MAEs, ViTs, and baseline FMs can be found in \cref{tab:fusion_modes}, in addition to \cref{fig:fusion_comparison}.

\begin{table*}[ht]
\centering
\footnotesize
\caption{\textbf{Evaluation of different multimodal and multitemporal fusion modes for MAEs, ViTs, and baseline FMs}. We report the weighted F1 score (\%) on TreeSatAI-TS and the mIoU (\%) on PASTIS-HD and FLAIR-HUB 20\%.}
\label{tab:fusion_modes}
\begin{tabular}{llllccccc}
\toprule
\multirow{2}{*}{\makecell{Model}} & \multirow{2}{*}{\makecell{Model size}} & \multirow{2}{*}{\makecell{Fusion mode}} & 
\multirow{2}{*}{\makecell{Modality groups}} &
\multirow{2}{*}{\makecell{\textbf{TreeSatAI-TS} \\ \;}} & 
\multirow{2}{*}{\makecell{\textbf{PASTIS-HD} \\ fold I}} & 
\multirow{2}{*}{\makecell{\textbf{FLAIR-HUB} \\ filt. 20\% / split 1}}  \\
 & & & & &  \\
\midrule
MAE & Base  & shared & & 76.1 & 66.1 & 62.5 \\
\rowcolor{black!5} MAE & Base & monotemp & & 77.0 & 66.0 & 62.4 \\ 
MAE & Base  & mod & & 78.4 & 68.9 & 63.3 \\ 
\rowcolor{black!5} MAE & Base & group & $\text{S1}_\text{ASC}$, $\text{S1}_\text{DES}$ & 78.5 & 68.8 & \bf{63.6} \\ 
MAE & Base & group & $\text{S1}_\text{ASC}$, $\text{S1}_\text{DES}$, S2 & 78.1 & \bf{69.1} & 63.5 \\ 
\rowcolor{black!5} MAE & Base & group & $\text{S1}_\text{ASC}$, $\text{S1}_\text{DES}$, S2, VHR & 78.2 & 68.1 & 61.4 \\ 
MAE & Base & inter-group & $\text{S1}_\text{ASC}$, $\text{S1}_\text{DES}$ & \bf{78.8} & 68.6 & 63.5 \\ 
\midrule
ViT & Base & shared & & 72.6 & 64.1 & 59.3 \\ 
\rowcolor{black!5} ViT & Base & monotemp & & 73.0 & 64.4 & 58.9 \\ 
ViT & Base & mod & & 75.4 & \bf{64.6} & 59.0 \\ 
\rowcolor{black!5} ViT & Base & group & $\text{S1}_\text{ASC}$, $\text{S1}_\text{DES}$ & 75.7 & 64.6 & \bf{59.5} \\ 
ViT & Base & group & $\text{S1}_\text{ASC}$, $\text{S1}_\text{DES}$, S2 & \bf{76.0} & 64.2 & 59.1 \\ 
\rowcolor{black!5}ViT & Base & group & $\text{S1}_\text{ASC}$, $\text{S1}_\text{DES}$, S2, VHR & 75.9 & 64.0 & 56.4 \\ 
ViT & Base & inter-group & $\text{S1}_\text{ASC}$, $\text{S1}_\text{DES}$ & 75.6 & 64.5 & 58.8 \\ 
\midrule
DINO-v2 \cite{oquab2023} & Base & shared && 76.7  & 64.4 & 64.2  \\
\rowcolor{black!5} DINO-v2 \cite{oquab2023} & Base & monotemp && 76.4  & 63.7  & 63.9  \\
DINO-v2 sat. \cite{tolan2024} & Large & shared && 76.3  & 64.0  &  \textbf{64.5} \\
  \rowcolor{black!5} DINO-v2 sat. \cite{tolan2024} & Large & monotemp && 76.6 & 63.1  & 64.0 \\
DOFA \cite{xiong2024} & Base & shared && 76.1 & 62.9 & 62.8 \\
 \rowcolor{black!5} DOFA \cite{xiong2024} & Base & monotemp && 75.9  & 63.2  & 64.1  \\
CROMA \cite{fuller2023} & Base & late-croma & & 69.8 & 64.9 & 41.9 \\
\rowcolor{black!5} CROMA \cite{fuller2023} & Base & inter-croma & & 70.5 & 65.0 & 42.5 \\
Prithvi-EO-2.0 \cite{szwarcman2024} & Large & mod & & 75.6 & 66.2 & 42.5 \\
\rowcolor{black!5} SatMAE \cite{cong2022} & Large & mod & & \textbf{76.9} & \textbf{66.6} & 42.8 \\
\bottomrule
\end{tabular}
\end{table*}

\subsection{Multispectral Fusion/Target Normalization}
\label{sm:multispectral_comparison}

The detailed numbers of the evaluation of the different choices of multispectral fusion and target normalization with MAEs can be found in \cref{tab:multispectral_fraction,tab:multispectral_size}, in addition to \cref{fig:multispectral_comparison}. We also report results after probing evaluation in \cref{tab:multispectral_fraction_probe,tab:multispectral_size_probe}.

We observe a slight performance drop when using token-based fusion with patch-wise/patch-group-wise normalization compared to joint-token fusion with patch-group-wise normalization. This may be attributed to the combination of two factors: (i) certain Sentinel-2 band groups may be less relevant to the downstream tasks, and (ii) our classification/segmentation heads, which aggregate multimodal and multitemporal features through a single attentive pooling layer, are shallow and may struggle to filter out irrelevant tokens—thereby introducing noise into the final predictions.

\begin{table*}[ht]
\centering
\footnotesize
\caption{\textbf{Evaluation of different choices of multispectral fusion and target normalization for MAE-B models across pre-training dataset fraction}. We report the weighted F1 score (\%) on TreeSatAI-TS and the mIoU (\%) on PASTIS-HD.}
\label{tab:multispectral_fraction}
\begin{tabular}{llllllll}
\toprule
\multirow{3}{*}{\makecell{Multispectral \\ fusion}} & 
\multirow{3}{*}{\makecell{Target \\ normalization}} & 
\multicolumn{3}{c}{\makecell{\textbf{TreeSatAI-TS} \\ \;}} &
\multicolumn{3}{c}{\makecell{\textbf{PASTIS-HD} \\ fold I}} \\
\cmidrule(lr){3-5} \cmidrule(lr){6-8}
& & \multicolumn{3}{c}{Pre-training dataset frac.} & \multicolumn{3}{c}{Pre-training dataset frac.} \\
& & 5\% & 20\% & 100\% & 5\% & 20\% & 100\% \\
\midrule
Joint-token & No normalization      & 72.7 & 72.8 & 74.6 & 64.9 & 65.4 & 65.1 \\
\rowcolor{black!5} Joint-token & Patch-wise            & 74.4 & 75.5 & 77.4 & 66.7 & 67.1 & 67.5 \\
Joint-token & Patch-group-wise      
 & \bf{76.3}
 & \bf{77.4}
 & \bf{78.5}
 & \bf{67.3}
 & \bf{68.1}
 & \bf{68.8} \\
\midrule
Token-based  & \makecell{Patch-wise/\\Patch-group-wise} & 76.6 & 77.0 & 78.9 & 66.7 & 67.2 & 68.6 \\
\bottomrule
\end{tabular}
\end{table*}

\begin{table*}[ht]
\centering
\footnotesize
\caption{\textbf{Evaluation of different choices of multispectral fusion and target normalization for MAE models across model size}. We report the weighted F1 score (\%) on TreeSatAI-TS and the mIoU (\%) on PASTIS-HD.}
\label{tab:multispectral_size}
\begin{tabular}{llllllll}
\toprule
\multirow{3}{*}{\makecell{Multispectral \\ fusion}} & 
\multirow{3}{*}{\makecell{Target \\ normalization}} & 
\multicolumn{3}{c}{\makecell{\textbf{TreeSatAI-TS} \\ \;}} &
\multicolumn{3}{c}{\makecell{\textbf{PASTIS-HD} \\ fold I}} \\
\cmidrule(lr){3-5} \cmidrule(lr){6-8}
& & \multicolumn{3}{c}{Model size} & \multicolumn{3}{c}{Model size} \\
& & Small & Base & Large & Small & Base & Large \\
\midrule
Joint-token & No normalization      
 & 74.9 & 74.6 & 74.9 
 & 64.9 & 65.1 & 65.0 \\
\rowcolor{black!5} Joint-token & Patch-wise            
 & 77.2 & 77.4 & 77.1 
 & 66.8 & 67.5 & 67.7 \\
Joint-token & Patch-group-wise      
 & \bf{77.3} & \bf{78.5} & \bf{78.5} 
 & \bf{67.8} & \bf{68.8} & \bf{69.2} \\
\midrule
Token-based & \makecell{Patch-wise/\\Patch-group-wise}         
 & 76.9 & 78.9 & 79.0 
 & 67.0 & 68.6 & 69.5 \\
\bottomrule
\end{tabular}
\end{table*}

\begin{table*}[ht]
\centering
\footnotesize
\caption{\textbf{Probing evaluation of different choices of multispectral fusion and target normalization for MAE-B models across pre-training dataset fraction}. We report the weighted F1 score (\%) on TreeSatAI-TS and the mIoU (\%) on PASTIS-HD.}
\label{tab:multispectral_fraction_probe}
\begin{tabular}{llllllll}
\toprule
\multirow{3}{*}{\makecell{Multispectral \\ fusion}} & 
\multirow{3}{*}{\makecell{Target \\ normalization}} & 
\multicolumn{3}{c}{\makecell{\textbf{TreeSatAI-TS} \\ \;}} &
\multicolumn{3}{c}{\makecell{\textbf{PASTIS-HD} \\ fold I}} \\
\cmidrule(lr){3-5} \cmidrule(lr){6-8}
& & \multicolumn{3}{c}{Pre-training dataset frac.} & \multicolumn{3}{c}{Pre-training dataset frac.} \\
& & 5\% & 20\% & 100\% & 5\% & 20\% & 100\% \\
\midrule
Joint-token & No normalization         & 57.2 & 61.1 & 63.2 & 52.5 & 56.9 & 57.9 \\
\rowcolor{black!5} Joint-token       & Patch-wise   & 62.9 & 65.1 & 67.7 & 58.2 & 59.7 & 61.2 \\
Joint-token       & Patch-group-wise   
 & \bf{64.8}
 & \bf{68.6}
 & \bf{69.3}
 & \bf{59.0}
 & \bf{60.2}
 & \bf{61.2}\\
\midrule
Token-based       & \makecell{Patch-wise/\\Patch-group-wise}         & 61.9 & 66.2 & 69.3 & 55.9 & 57.5 & 61.0 \\
\bottomrule
\end{tabular}
\end{table*}

\begin{table*}[ht]
\centering
\footnotesize
\caption{\textbf{Probing evaluation of different choices of multispectral fusion and target normalization for MAE models across model size}. We report the weighted F1 score (\%) on TreeSatAI-TS and the mIoU (\%) on PASTIS-HD.}
\label{tab:multispectral_size_probe}
\begin{tabular}{llllllll}
\toprule
\multirow{3}{*}{\makecell{Multispectral \\ fusion}} & 
\multirow{3}{*}{\makecell{Target \\ normalization}} & 
\multicolumn{3}{c}{\makecell{\textbf{TreeSatAI-TS} \\ \;}} &
\multicolumn{3}{c}{\makecell{\textbf{PASTIS-HD} \\ fold I}} \\
\cmidrule(lr){3-5} \cmidrule(lr){6-8}
& & \multicolumn{3}{c}{Model size} & \multicolumn{3}{c}{Model size} \\
& & Small & Base & Large & Small & Base & Large \\
\midrule
Joint-token & No normalization      
 & 57.3 & 63.2 & 64.1 
 & 54.1 & 57.9 & 58.3 \\
\rowcolor{black!5} Joint-token & Patch-wise            
 & 63.1 & 67.7 & 68.4 
 &  \textbf{57.6} & 61.2 & 62.1 \\
Joint-token & Patch-group-wise      
 &  \textbf{64.3} &  \textbf{69.3} &  \textbf{71.4} 
 & 55.1 & \textbf{61.2} &  \textbf{63.6} \\
\midrule
Token-based & \makecell{Patch-wise/\\Patch-group-wise}          
 & 61.7 & 69.3 & 69.4 
 & 53.8 & 61.0 & 63.9 \\
\bottomrule
\end{tabular}
\end{table*}

\subsection{Scaling by Dataset Size}
\label{sm:Scaling_byDataset_Size}

The detailed numbers of the performance with different pre-training/fine-tuning dataset fractions for MAE-B and ViT-B models are provided in \cref{tab:mae_pretrain_finetune}, in addition to \cref{fig:scaling}.

\begin{table*}[ht]
\centering
\footnotesize
\caption{\textbf{Scaling of MAE-B and ViT-B models with different pre-training/fine-tuning dataset fractions.} We report the weighted F1 score (\%) on TreeSatAI-TS and the mIoU (\%) on PASTIS-HD and FLAIR-HUB for three fine-tuning dataset fractions: 5\%, 20\%, and 100\%. For each fine-tuning fraction, we compare three pre-training settings: no pre-training, pre-training on the same fraction as fine-tuning, and pre-training on 100\% of the data. Note that pre-training on the same fraction as fine-tuning and pre-training on 100\% of the data become equivalent when fine-tuning on 100\% of the data, hence the identical performance.}
\label{tab:mae_pretrain_finetune}
\begin{tabular}{lccccccccc}
\toprule
& \multicolumn{3}{c}{\makecell{\textbf{TreeSatAI-TS} \\ \;}} & 
\multicolumn{3}{c}{\makecell{\textbf{PASTIS-HD} \\ fold I}} & 
\multicolumn{3}{c}{\makecell{\textbf{FLAIR-HUB} \\ split 1}} \\
\cmidrule(lr){2-4} \cmidrule(lr){5-7} \cmidrule(lr){8-10}
& \multicolumn{3}{c}{Fine-tuning dataset frac.} & \multicolumn{3}{c}{Fine-tuning dataset frac.} & \multicolumn{3}{c}{Fine-tuning dataset frac.} \\
& 5\% & 20\% & 100\% & 5\% & 20\% & 100\% & 5\% & 20\% & 100\% \\
\midrule
No pre-training & 61.8 & 69.0 & 75.7 & 38.8 & 52.2 & 64.6 & 55.3 & 59.5 & 61.6 \\
\rowcolor{black!5} Pre-training fraction =  Fine-tuning fraction & 65.5 & 71.8 & 78.5 & 46.6 & 57.1 & 68.8 & 60.1 & 63.6 & 64.9 \\
Pre-training fraction = 100\% & \textbf{67.8} & \textbf{73.8} & \textbf{78.5} & \textbf{52.5} & \textbf{59.2} & \textbf{68.8} & \textbf{61.5} & \textbf{64.6} & \textbf{64.9} \\
\bottomrule
\end{tabular}
\end{table*}

%% file: 8_add_results.tex
\section{Additional Results}
\label{sm:add_results}

In this section, we report additional ablations on our masking strategy (\cref{sm:masking_comparison}), our choices of encodings (\cref{sm:ablation_encodings}), the impact of multitemporal components (\cref{sm:impact_multitemporal}), and the importance of each modality by dataset (\cref{sm:impact_modality}).

\subsection{Masking Strategy}
\label{sm:masking_comparison}

We examine the impact of various components of our masking strategy. Specifically, we assess the effect of including the different structured masking steps—namely steps (i-a), (i-b), and (i-c) detailed in \cref{sm:masking}—in the SSL pretext task. In these experiments, we use the \textit{group} fusion mode, grouping together the Sentinel-1 ascending and descending modalities. For multispectral data, we use \textit{joint-token} fusion combined with \textit{patch-group-wise} target normalization during reconstruction.

We report results in \cref{tab:masking_modes_probe}. We also report results after probing evaluation in \cref{tab:masking_modes}. 

We find that structured masking yields only marginal improvements in fine-tuning performance. However, temporally structured masking leads to a notable boost in probing performance on PASTIS-HD. This suggests that structured masking may produce more useful representations when evaluated directly (i.e., without fine-tuning), although these benefits do not necessarily transfer fully after model fine-tuning.

\begin{table}[ht]
\centering
\footnotesize
\caption{\textbf{Evaluation of different choices of masking for MAE-B models}. We report the weighted F1 score (\%) on TreeSatAI-TS and the mIoU (\%) on PASTIS-HD and FLAIR-HUB 20\%. Arrows ($\uparrow$ / $\downarrow$) indicate the difference compared to our default masking strategy with modality, spatial and temporal structure.}
\label{tab:masking_modes}
\begin{tabular}{cccccc}
\toprule
\multicolumn{3}{c}{Masking structure} & 
\multirow{2.6}{*}{\makecell{\textbf{TreeSatAI-TS}\\ \;}} & 
\multirow{2.6}{*}{\makecell{\textbf{PASTIS-HD}\\ fold I}} & 
\multirow{2.6}{*}{\makecell{\textbf{FLAIR-HUB}\\ filt. 20\% / split 1}} \\
\cmidrule(lr){1-3}
Modality & Spatial & Temporal & \\
\midrule
\cmark & \cmark & \cmark & \textbf{78.5} \makebox[0pt][l]{\,} & \textbf{68.8} \makebox[0pt][l]{\,} & \textbf{63.6} \makebox[0pt][l]{\,} \\
\rowcolor{black!5} \xmark & \cmark & \cmark & 78.5 \makebox[0pt][l]{\scriptsize\textcolor{Green}{$\uparrow$0.0}} & 68.4 \makebox[0pt][l]{\scriptsize\textcolor{Red}{$\downarrow$0.4}} & 63.3 \makebox[0pt][l]{\scriptsize\textcolor{Red}{$\downarrow$0.3}} \\ 
\cmark & \xmark & \cmark & 78.5 \makebox[0pt][l]{\scriptsize\textcolor{Green}{$\uparrow$0.0}} & 68.6 \makebox[0pt][l]{\scriptsize\textcolor{Red}{$\downarrow$0.2}} & 63.0  \makebox[0pt][l]{\scriptsize\textcolor{Red}{$\downarrow$0.6}} \\ 
\rowcolor{black!5} \cmark & \cmark & \xmark & 78.3 \makebox[0pt][l]{\scriptsize\textcolor{Red}{$\downarrow$0.2}} & 68.5 \makebox[0pt][l]{\scriptsize\textcolor{Red}{$\downarrow$0.3}} & 63.4 \makebox[0pt][l]{\scriptsize\textcolor{Red}{$\downarrow$0.2}} \\ 
\xmark & \xmark & \xmark & 78.2 \makebox[0pt][l]{\scriptsize\textcolor{Red}{$\downarrow$0.3}} & 68.4 \makebox[0pt][l]{\scriptsize\textcolor{Red}{$\downarrow$0.4}} & 63.6  \makebox[0pt][l]{\scriptsize\textcolor{Green}{$\uparrow$0.0}} \\ 
\bottomrule
\end{tabular}
\end{table}

\begin{table}[ht]
\centering
\footnotesize
\caption{\textbf{Probing evaluation of different choices of masking for MAE-B models}. We report the weighted F1 score (\%) on TreeSatAI-TS and the mIoU (\%) on PASTIS-HD and FLAIR-HUB 20\%. Arrows ($\uparrow$ / $\downarrow$) indicate the difference compared to our default masking strategy with modality, spatial and temporal structure.}
\label{tab:masking_modes_probe}
\begin{tabular}{cccccc}
\toprule
\multicolumn{3}{c}{Masking structure} & 
\multirow{2.6}{*}{\makecell{\textbf{TreeSatAI-TS}\\ \;}} & 
\multirow{2.6}{*}{\makecell{\textbf{PASTIS-HD}\\ fold I}} & 
\multirow{2.6}{*}{\makecell{\textbf{FLAIR-HUB}\\ filt. 20\% / split 1}} \\
\cmidrule(lr){1-3}
Modality & Spatial & Temporal &  &  &  \\
\midrule
\cmark & \cmark & \cmark & 69.3 \makebox[0pt][l]{\,} & 61.2 \makebox[0pt][l]{\,} & 56.2 \makebox[0pt][l]{\,} \\
\rowcolor{black!5} \xmark & \cmark & \cmark & 69.8 \makebox[0pt][l]{\scriptsize\textcolor{Green}{$\uparrow$0.5}} & 61.4 \makebox[0pt][l]{\scriptsize\textcolor{Green}{$\uparrow$0.2}} & 56.3 \makebox[0pt][l]{\scriptsize\textcolor{Green}{$\uparrow$0.1}} \\
\cmark & \xmark & \cmark & 69.6  \makebox[0pt][l]{\scriptsize\textcolor{Green}{$\uparrow$0.3}} & \textbf{61.5}  \makebox[0pt][l]{\scriptsize\textcolor{Green}{$\uparrow$0.3}} & 56.3 \makebox[0pt][l]{\scriptsize\textcolor{Green}{$\uparrow$0.1}} \\
\rowcolor{black!5} \cmark & \cmark & \xmark & 69.2  \makebox[0pt][l]{\scriptsize\textcolor{Red}{$\downarrow$0.1}} & 57.7 \makebox[0pt][l]{\scriptsize\textcolor{Red}{$\downarrow$3.5}} & 56.2 \makebox[0pt][l]{\scriptsize\textcolor{Green}{$\uparrow$0.0}} \\
\xmark & \xmark & \xmark & \textbf{69.9} \makebox[0pt][l]{\scriptsize\textcolor{Green}{$\uparrow$0.6}} & 57.6 \makebox[0pt][l]{\scriptsize\textcolor{Red}{$\downarrow$3.6}} & \textbf{56.3} \makebox[0pt][l]{\scriptsize\textcolor{Green}{$\uparrow$0.1}} \\
\bottomrule
\end{tabular}
\end{table}

\subsection{Ablation of Temporal/Modality Encodings}
\label{sm:ablation_encodings}

We evaluate the effectiveness of the encoding strategies described in \cref{sm:encodings}. In these experiments, we use the \textit{group} fusion mode, grouping together the Sentinel-1 ascending and descending modalities. For multispectral data, we use \textit{joint-token} fusion combined with \textit{patch-group-wise} target normalization during reconstruction.

First, we evaluate the impact of removing temporal encodings. In this case, positional encodings occupy the full $C_e$ dimensions, and aggregated encodings are formed solely from them.

As shown in \cref{tab:ablation_encodings}, including these temporal encodings is critical on tasks which are heavily tied to multitemporal dynamics, such as TreeSatAI-TS and PASTIS-HD. Such temporal encodings constitute the only mechanism allowing the model to contextualize tokens across time steps. Without them, the model \say{sees} multiple dates but must infer their temporal positions and ordering. This ambiguity is especially detrimental for tasks requiring precise multitemporal signatures, such as distinguishing tree species’ phenologies in TreeSatAI-TS or crop types in PASTIS-HD.

Next, we investigate the addition of modality encodings, implemented as learnable modality-specific parameters of dimension $8$. When including these modality encodings, we reduce the positional encoding dimensionality to $C_e-16$ and form the aggregated encodings by concatenating the modality, temporal, and positional encodings of dimensions $8$, $8$, and $C_e-16$, respectively.

As shown in \cref{tab:ablation_encodings}, including explicit modality encodings has only a slight and inconsistent effect. This confirms that modality-specific tokenizers and learnable modality-specific \verb|[mask]| tokens already provide sufficient modality differentiation. The small performance drop observed on FLAIR-HUB 20\% may stem from the reduction in positional encoding dimensionality.

\begin{table}[ht!]
\centering
\footnotesize
\caption{\textbf{Ablation of temporal and modality encodings for MAE-B and ViT-B models.} We report the weighted F1 score (\%) on TreeSatAI-TS and the mIoU (\%) on PASTIS-HD and FLAIR-HUB 20\%. Arrows ($\uparrow$ / $\downarrow$)  indicate the difference compared to using temporal encodings but not modality encodings.}
\label{tab:ablation_encodings}
\begin{tabular}{lccccc}
\toprule
Model & 
Temporal enc. & 
Modality enc. & 
\makecell{\textbf{TreeSatAI-TS} \\ \,} & 
\makecell{\textbf{PASTIS-HD}\\ fold I} & 
\makecell{\textbf{FLAIR-HUB}\\ filt. 20\% / split 1} \\
\midrule
MAE & \cmark & \xmark & 78.5 \makebox[0pt][l]{\,} & 68.8 \makebox[0pt][l]{\,} & \textbf{63.6} \makebox[0pt][l]{\,} \\
\rowcolor{black!5} MAE & \xmark & \xmark & 77.8 \makebox[0pt][l]{\scriptsize\color{Red}$\downarrow$0.7} & 67.7 \makebox[0pt][l]{\scriptsize\color{Red}$\downarrow$1.1} & 63.3 \makebox[0pt][l]{\scriptsize\color{Red}$\downarrow$0.3} \\
MAE & \cmark & \cmark & \textbf{78.7} \makebox[0pt][l]{\scriptsize\color{Green}$\uparrow$0.2} & \textbf{69.0} \makebox[0pt][l]{\scriptsize\color{Green}$\uparrow$0.2} & 63.2 \makebox[0pt][l]{\scriptsize\color{Red}$\downarrow$0.4} \\
\midrule
ViT & \cmark & \xmark & \textbf{75.7} \makebox[0pt][l]{\,} & \textbf{64.6} \makebox[0pt][l]{\,} & \textbf{59.5} \makebox[0pt][l]{\,} \\
\rowcolor{black!5} ViT & \xmark & \xmark & 74.7 \makebox[0pt][l]{\scriptsize\color{Red}$\downarrow$1.0} & 63.6 \makebox[0pt][l]{\scriptsize\color{Red}$\downarrow$1.0} & 59.2 \makebox[0pt][l]{\scriptsize\color{Red}$\downarrow$0.3} \\
ViT & \cmark & \cmark & 75.6 \makebox[0pt][l]{\scriptsize\color{Red}$\downarrow$0.1} & 64.6 \makebox[0pt][l]{\scriptsize\color{Green}$\uparrow$0.0} & 58.9 \makebox[0pt][l]{\scriptsize\color{Red}$\downarrow$0.6} \\
\bottomrule
\end{tabular}
\end{table}

\subsection{Importance of Multitemporal Components}
\label{sm:impact_multitemporal}

We examine the impact of various multitemporal components on downstream task performance for MAE-B and ViT-B models. In these experiments, we use the \textit{group} fusion mode, grouping together the Sentinel-1 ascending and descending modalities. For multispectral data, we use \textit{joint-token} fusion combined with \textit{patch-group-wise} target normalization during reconstruction.

First, we evaluate the effect of varying the number of temporal bins $D_m$ for the Sentinel-2 and Sentinel-1 ascending and descending modalities. As shown in \cref{tab:impact_timesteps}, the number of temporal bins $D_m$ for Sentinel-2 is critical when this modality drives performance, as in TreeSatAI-TS and PASTIS-HD (see \cref{sm:impact_modality}). Notably, the performance gap between $D_m=16$ and $D_m=1$ for Sentinel-2 is much larger than that between $D_m=1$ and $D_m=0$. This indicates that it is the multitemporal dynamics of Sentinel-2, rather than its monotemporal component, that is essential. In contrast, varying $D_m$ for Sentinel-1 has only a minor effect, reflecting its smaller role (see \cref{sm:impact_modality}).

\begin{table*}[ht]
\centering
\footnotesize
\caption{\textbf{Importance of the number of temporal bins for MAE-B and ViT-B models.} We report the weighted F1 score (\%) on TreeSatAI-TS and the mIoU (\%) on PASTIS-HD and FLAIR-HUB 20\%. Arrows ($\uparrow$ / $\downarrow$)  indicate the difference compared to using 16 temporal bins for the Sentinel-2 modality and 4 temporal bins for the Sentinel-1 ascending and descending modalities.}
\label{tab:impact_timesteps}

\begin{tabular}{l 
>{\centering\arraybackslash}p{.7cm} 
>{\centering\arraybackslash}p{.7cm} 
cccc}
\toprule
\multirow{2.6}{*}{Model} & 
\multicolumn{3}{c}{\makecell{Number of temporal bins}} & 
\multirow{2.6}{*}{\makecell{\textbf{TreeSatAI-TS}\\ \;}} & 
\multirow{2.6}{*}{\makecell{\textbf{PASTIS-HD}\\ fold I}} & 
\multirow{2.6}{*}{\makecell{\textbf{FLAIR-HUB}\\ filt. 20\% / split 1}} \\
\cmidrule(lr){2-4}
 & \(\text{S1}_\text{ASC}\) & \(\text{S1}_\text{DES}\) & S2 &  & \\
\midrule
					MAE & 4  & 4 & 16 & \textbf{78.5} \makebox[0pt][l]{\,} & 68.8 \makebox[0pt][l]{\,} & \textbf{63.6} \makebox[0pt][l]{\,} \\
\rowcolor{black!5}  MAE & 4  & 4 & 4 & 75.3 \makebox[0pt][l]{\scriptsize\color{Red} $\downarrow$3.2} & 64.3 \makebox[0pt][l]{\scriptsize\color{Red}$\downarrow$4.5} & 62.4 \makebox[0pt][l]{\scriptsize\color{Red}$\downarrow$1.2} \\
 					MAE & 4  & 4 & 1 & 73.1 \makebox[0pt][l]{\scriptsize\color{Red} $\downarrow$5.4} & 49.3 \makebox[0pt][l]{\scriptsize\color{Red}$\downarrow$19.5} & 61.8 \makebox[0pt][l]{\scriptsize\color{Red}$\downarrow$1.8} \\
\rowcolor{black!5} 	MAE & 4  & 4 & 0 & 72.6 \makebox[0pt][l]{\bf\scriptsize\color{Red} $\downarrow$5.9} & 44.9 \makebox[0pt][l]{\bf\scriptsize\color{Red}$\downarrow$23.9} & 61.6 \makebox[0pt][l]{\bf\scriptsize\color{Red}$\downarrow$2.0} \\
 			 	    MAE & 1  & 1 & 16 & 78.3 \makebox[0pt][l]{\scriptsize\color{Red} $\downarrow$0.2} & 68.8 \makebox[0pt][l]{\scriptsize\color{Green}$\uparrow$0.0} & 63.5 \makebox[0pt][l]{\scriptsize\color{Red}$\downarrow$0.1} \\
\rowcolor{black!5}	MAE & 0 & 0 & 16 & 78.2 \makebox[0pt][l]{\scriptsize\color{Red} $\downarrow$0.3} & \textbf{69.2} \makebox[0pt][l]{\scriptsize\color{Green}$\uparrow$0.4} & 63.1 \makebox[0pt][l]{\scriptsize\color{Red}$\downarrow$0.5} \\
\midrule
					ViT & 4  & 4 & 16 & 75.7 \makebox[0pt][l]{\,} & \textbf{64.6} \makebox[0pt][l]{\,} & 59.5 \makebox[0pt][l]{\,} \\
\rowcolor{black!5}  ViT & 4  & 4 & 4 & 70.6 \makebox[0pt][l]{\scriptsize\color{Red} $\downarrow$5.1} & 61.2 \makebox[0pt][l]{\scriptsize\color{Red}$\downarrow$3.4} & 58.4 \makebox[0pt][l]{\scriptsize\color{Red}$\downarrow$1.1} \\
					ViT & 4  & 4 & 1 & 66.1 \makebox[0pt][l]{\scriptsize\color{Red} $\downarrow$9.6} & 46.5 \makebox[0pt][l]{\scriptsize\color{Red}$\downarrow$18.1} & 57.6 \makebox[0pt][l]{\scriptsize\color{Red}$\downarrow$1.9} \\
\rowcolor{black!5} 	ViT & 4  & 4 & 0 & 65.8 \makebox[0pt][l]{\bf\scriptsize\color{Red} $\downarrow$9.9} & 41.5 \makebox[0pt][l]
{\bf\scriptsize\color{Red}$\downarrow$23.1} & 57.4 \makebox[0pt][l]{\bf\scriptsize\color{Red}$\downarrow$2.1} \\
				    ViT & 1 & 1 & 16 & \textbf{75.8} \makebox[0pt][l]{\scriptsize\color{Green} $\uparrow$0.1} & 64.6 \makebox[0pt][l]{\scriptsize\color{Green}$\uparrow$0.0} & \textbf{59.8} \makebox[0pt][l]{\scriptsize\color{Green}$\uparrow$0.3} \\
\rowcolor{black!5}  ViT & 0 & 0 & 16 & 75.7 \makebox[0pt][l]{\scriptsize\color{Green} $\uparrow$0.0} & 64.4 \makebox[0pt][l]{\scriptsize\color{Red}$\downarrow$0.2} & 59.2 \makebox[0pt][l]{\scriptsize\color{Red}$\downarrow$0.3} \\
\bottomrule
\end{tabular}
\end{table*}

Next, we assess the impact of disabling the data augmentation associated with random time step selection within temporal bins. As shown in \cref{tab:impact_random_selection}, disabling this data augmentation generally degrades performance, although the magnitude of the degradation varies across datasets.

\begin{table*}[ht]
\centering
\footnotesize
\caption{\textbf{Importance of random time step selection for MAE-B and ViT-B models.} We report the weighted F1 score (\%) on TreeSatAI-TS and the mIoU (\%) on PASTIS-HD and FLAIR-HUB 20\%. Arrows ($\uparrow$ / $\downarrow$)  indicate the difference compared to enabling random time step selection.}
\label{tab:impact_random_selection}

\begin{tabular}{l 
>{\centering\arraybackslash}p{.7cm} 
>{\centering\arraybackslash}p{.7cm} 
>{\centering\arraybackslash}p{.7cm} 
cccc}
\toprule
\multirow{2.6}{*}{Model} & 
\multicolumn{3}{c}{\makecell{Number of temporal bins}} & 
\multirow{2.6}{*}{\makecell{Random time step\\ selection}} &
\multirow{2.6}{*}{\makecell{\textbf{TreeSatAI-TS}\\ \;}} & 
\multirow{2.6}{*}{\makecell{\textbf{PASTIS-HD}\\ fold I}} & 
\multirow{2.6}{*}{\makecell{\textbf{FLAIR-HUB}\\ filt. 20\% / split 1}} \\
\cmidrule(lr){2-4}
 & \(\text{S1}_\text{ASC}\) & \(\text{S1}_\text{DES}\) & S2 & &  & \\
\midrule
					MAE & 4  & 4 & 16 & \cmark & 78.5 \makebox[0pt][l]{\,} & 68.8 \makebox[0pt][l]{\,} & 63.6 \makebox[0pt][l]{\,} \\
\rowcolor{black!5}  MAE & 4  & 4 & 16 & \xmark & \textbf{78.8} \makebox[0pt][l]{\scriptsize\color{Green}$\uparrow$0.3} & 68.2 \makebox[0pt][l]{\scriptsize\color{Red}$\downarrow$0.6} & \textbf{63.7} \makebox[0pt][l]{\scriptsize\color{Green}$\uparrow$0.1} \\
\midrule
					ViT & 4  & 4 & 16 & \cmark & \textbf{75.7} \makebox[0pt][l]{\,} & \textbf{64.6} \makebox[0pt][l]{\,} & \textbf{59.5} \makebox[0pt][l]{\,} \\
\rowcolor{black!5}  ViT & 4  & 4 & 16 & \xmark & 75.4 \makebox[0pt][l]{\scriptsize\color{Red}$\downarrow$0.3} & 63.9 \makebox[0pt][l]{\scriptsize\color{Red}$\downarrow$0.7} & 58.7 \makebox[0pt][l]{\scriptsize\color{Red}$\downarrow$0.8} \\
\bottomrule
\end{tabular}
\end{table*}

\subsection{Importance of each Modality by Dataset}
\label{sm:impact_modality}

We present an ablation study on modality removal for MAE-B and ViT-B models. In these experiments, we use the \textit{group} fusion mode, grouping together the Sentinel-1 ascending and descending modalities. For multispectral data, we use \textit{joint-token} fusion along with \textit{patch-group-wise} target normalization during reconstruction.

As shown in \cref{tab:ablation_modalities}, the best performance is generally obtained when all modalities are included. The Sentinel-2 modality has a strong influence on the TreeSatAI-TS and PASTIS-HD tasks, which are heavily tied to multitemporal dynamics. Additionally, the modality from which the ground truth is derived on FLAIR-HUB (aerial) markedly affects performance.

The slight performance drop observed when omitting Sentinel-1 on PASTIS-HD may stem from the combination of two factors: (i) its lower task relevance compared to Sentinel-2, and (ii) the relative shallowness of segmentation heads, which may struggle to filter out irrelevant tokens—thereby introducing noise into the final predictions.

\begin{table*}[ht]
\centering
\footnotesize
\caption{\textbf{Ablation study on modality removal for MAE-B and ViT-B models}. We report the weighted F1 score (\%) on TreeSatAI-TS and the mIoU (\%) on PASTIS-HD and FLAIR-HUB 20\%. Arrows ($\uparrow$ / $\downarrow$)  indicate the difference compared to including all modalities.}
\label{tab:ablation_modalities}

\begin{tabular}{l ccc l ccc l cccc l}
\toprule
\multirow{2.6}{*}{Model} & 
\multicolumn{4}{c}{\makecell{\textbf{TreeSatAI-TS} \\ \;}} &
\multicolumn{4}{c}{\makecell{\textbf{PASTIS-HD} \\ fold I}} &
\multicolumn{5}{c}{\makecell{\textbf{FLAIR-HUB} \\ filt. 20\% / split 1}} \\
\cmidrule(lr){2-5} \cmidrule(lr){6-9} \cmidrule(lr){10-14}
 & Aerial & S1 & S2 & wF1 (\%) & Spot & S1 & S2 & mIoU (\%) & Aerial & S1 & S2 & DEM/DSM & mIoU (\%)  \\
\midrule
MAE
 & \cmark & \cmark & \cmark & \textbf{78.5}
 & \cmark & \cmark & \cmark & 68.8
 & \cmark & \cmark & \cmark & \cmark & \textbf{63.6} \\
\rowcolor{black!5} MAE
 & \xmark & \cmark & \cmark & 76.8 {\scriptsize\color{Red}$\downarrow$1.7}
 & \xmark & \cmark & \cmark & 68.6 {\scriptsize\color{Red}$\downarrow$0.2}
 & \xmark & \cmark & \cmark & \cmark & 52.2 {\scriptsize\color{Red}$\downarrow$\textbf{11.4}} \\
MAE & \cmark & \xmark & \cmark & 78.2 {\scriptsize\color{Red}$\downarrow$0.3}
 & \cmark & \xmark & \cmark & \textbf{69.2} {\scriptsize\color{Green}$\uparrow$0.4}
 & \cmark & \xmark & \cmark & \cmark & 63.1 {\scriptsize\color{Red}$\downarrow$0.5} \\
\rowcolor{black!5} MAE
 & \cmark & \cmark & \xmark & 72.6 {\scriptsize\color{Red}$\downarrow$\textbf{5.9}}
 & \cmark & \cmark & \xmark & 44.9 {\scriptsize\color{Red}$\downarrow$\textbf{23.9}}
 & \cmark & \cmark & \xmark & \cmark & 61.6 {\scriptsize\color{Red}$\downarrow$2.0} \\
MAE &       &        &        &
 &        &        &        &
 & \cmark & \cmark & \cmark & \xmark & 62.7 {\scriptsize\color{Red}$\downarrow$0.9} \\
\midrule
ViT
 & \cmark & \cmark & \cmark & \textbf{75.7}
 & \cmark & \cmark & \cmark & \textbf{64.6}
 & \cmark & \cmark & \cmark & \cmark & \textbf{59.5} \\
\rowcolor{black!5} ViT
 & \xmark & \cmark & \cmark & 75.3 {\scriptsize\color{Red}$\downarrow$0.4}
 & \xmark & \cmark & \cmark & 64.5 {\scriptsize\color{Red}$\downarrow$0.1}
 & \xmark & \cmark & \cmark & \cmark & 47.9 {\scriptsize\color{Red}$\downarrow$\textbf{11.6}} \\
ViT & \cmark & \xmark & \cmark & 75.7 {\scriptsize\color{Green}$\uparrow$0.0}
 & \cmark & \xmark & \cmark & 64.4 {\scriptsize\color{Red}$\downarrow$0.2}
 & \cmark & \xmark & \cmark & \cmark & 59.2 {\scriptsize\color{Red}$\downarrow$0.3} \\
\rowcolor{black!5} ViT
 & \cmark & \cmark & \xmark & 65.8 {\scriptsize\color{Red}$\downarrow$\textbf{9.9}}
 & \cmark & \cmark & \xmark & 41.5 {\scriptsize\color{Red}$\downarrow$\textbf{23.1}}
 & \cmark & \cmark & \xmark & \cmark & 57.4 {\scriptsize\color{Red}$\downarrow$2.1} \\
ViT &       &        &        &
 &        &        &        &
 & \cmark & \cmark & \cmark & \xmark & 58.3 {\scriptsize\color{Red}$\downarrow$1.2} \\
\bottomrule
\end{tabular}
\end{table*}

%% file: 9_inference.tex
\section{Inference Results}
\label{sm:quali_results}

In this section, we report inference results from \emph{intra-dataset} MAE-B and supervised ViT-B models on the segmentation tasks of PASTIS-HD and FLAIR-HUB. We consider \emph{intra-dataset} MAESTRO and ViT models with the \textit{group} fusion mode, grouping together the Sentinel-1 ascending and descending modalities. For multispectral data, we use \textit{joint-token} fusion along with \textit{patch-group-wise} target normalization during reconstruction.

\begin{figure}[H]
    \centering
    \includegraphics[width=0.66\textwidth]{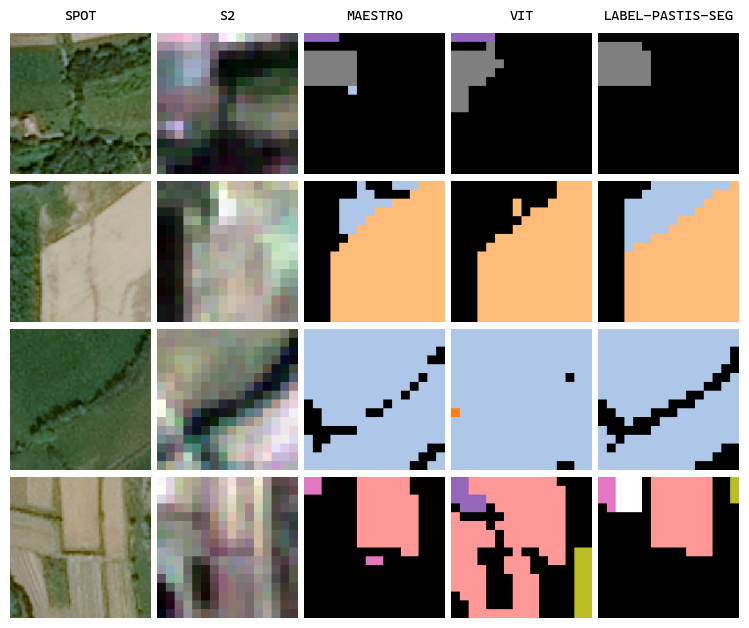}
	\caption{\textbf{Inference results from intra-dataset MAESTRO-B and ViT-B models on PASTIS-HD.} For Sentinel-2 imagery, we report the pixel-wise median across temporal bins. White parcels correspond to areas with missing annotations (void labels).}
    \label{fig:inferences_pastis}
\end{figure}

\begin{figure}[H]
    \centering
    \includegraphics[width=0.66\textwidth]{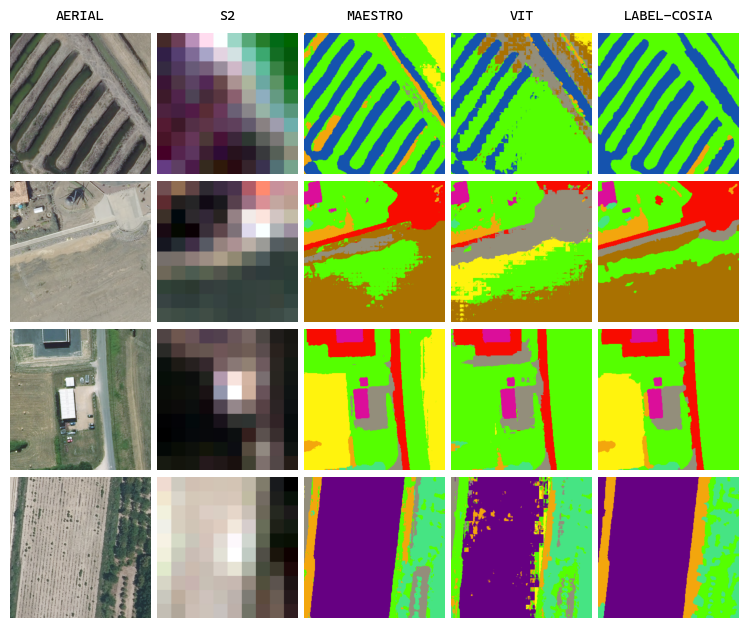}
    \caption{\textbf{Inference results from intra-dataset MAESTRO-B and ViT-B models on FLAIR-HUB.} For Sentinel-2 imagery, we report the pixel-wise median across temporal bins.}
    \label{fig:inferences_flair}
\end{figure}

Results are reported in \cref{fig:inferences_pastis} and \cref{fig:inferences_flair} for PASTIS-HD and FLAIR-HUB, respectively. For each tile, we display the VHR imagery, the Sentinel-2 imagery, the MAESTRO prediction, the ViT prediction, and the corresponding ground truth. Examples are randomly sampled from the test set.

In \cref{fig:inferences_pastis}, MAESTRO produces precise segmentation masks on PASTIS-HD, closely matching the boundaries of parcel and tree-covered areas. Its predictions are more spatially coherent than those of ViTs and better differentiate crop types.

In \cref{fig:inferences_flair}, although the complex scenes from FLAIR-HUB introduce prediction ambiguities, MAESTRO still delivers sharper and more accurate object delineations than ViTs. It also performs better on classes such as brushwood and water, and more effectively separates impervious from pervious surfaces.

%% file: 10_flops.tex
\section{Computational Costs in FLOPs}
\label{sm:flops}

In this section, we quantify floating-point operations (FLOPs) in our models.

We report FLOPs per forward pass (for a single batch element) separately for SSL pre-training (\cref{sm:macs_flops_pretraining}) and probing/fine-tuning (\cref{sm:macs_flops_probing_finetuning}). While forward FLOPs are identical between probing and fine-tuning, probing is substantially more efficient when backward FLOPs are taken into account.

We first calculate the number of multiply-accumulate operations (MACs), then convert to FLOPs using the relation $\text{FLOPs} = 2 \times \text{MACs}$. Following standard practice, we include only operations from model components with significant MAC contributions, excluding element-wise operations such as normalizations, nonlinearities, biases, and component-wise summations.

We consider the \emph{group} fusion mode, grouping the Sentinel-1 ascending and descending modalities together. For multispectral data, we evaluate both \emph{joint-token} and \emph{token-based} multispectral fusion.

\subsection{Pre-training FLOPs}
\label{sm:macs_flops_pretraining}

For each modality $m$, let $L_m$ denote its sequence length. With \emph{joint-token} multispectral fusion, the sequence length is $L_m = \left(I_m / P_m\right)^2 D_m$, where $I_m$ is the image size, $P_m$ the patch size, and $D_m$ the number of temporal bins. With \emph{token-based} multispectral fusion, the sequence length becomes $L_m = \left(I_m / P_m \right)^2 D_m |\mathcal{G}_m|$, where $|\mathcal{G}_m|$ is the cardinality of the set of band groups mapped to different tokens.

For a modality $m$ that is not grouped with others via early multimodal fusion, the MACs in the \emph{encoder} and \emph{decoder} are:
\begin{align*}
\text{MACs}^\text{enc} & = \left( 12 \lfloor (1-M) L_m \rceil C_e^2 + 2 \lfloor (1-M)  L_m \rceil^2 C_e \right) N_e, \\
\text{MACs}^\text{dec} & = \left( 12 L_m C_d^2 + 2 L_m^2 C_d \right) N_d,
\end{align*}
where $M$ is the fraction of masked tokens, $N_e$ and $N_d$ are the number of layers in the encoder and decoder, while $C_e$ and $C_d$ are the latent dimensions in the encoder and decoder.

For the grouped Sentinel-1 ascending and descending modalities, $\text{MACs}^\text{enc}$ and $\text{MACs}^\text{dec}$ are computed by replacing $L_m$ with the \textit{sum of sequence lengths} of the two modalities.

The \emph{dense} layers that project from the encoder space to the decoder space contribute:
\begin{align*}
\text{MACs}^\text{enc-to-dec} = \sum_m \lfloor (1-M) L_m \rceil C_e C_d.
\end{align*}

The \emph{patchify} and \emph{unpatchify} operations contribute:
\begin{align*}
\text{MACs}^\text{patchify} & = \sum_m I_m^2  D_m C_m C_e, \\
\text{MACs}^\text{unpatchify} & = \sum_m I_m^2 D_m C_m C_d,
\end{align*}
where $C_m$ is the number of channels for each modality $m$.

Putting everything together, we compute in \cref{tab:macs_flops_pretraining} the forward FLOPs ($\text{FLOPs} = 2 \times \text{MACs}$) for MAE models during pre-training. FLOPs are computed for both \emph{joint-token} and \emph{token-based} multispectral fusion across three model sizes and the four evaluated datasets. 

The computation is based on the default configurations in \cref{tab:treesat_hyperparams}, \cref{tab:pastis_hyperparams}, and \cref{tab:flair_hyperparams} and the values $M=0.75$ for the fraction of masked tokens, and $N_e=\{12,12,24\}$, $C_e=\{384,768,1024\}$, $N_d=\{2,3,4\}$, and $C_d=\{512,512,512\}$ for model sizes $\{\text{Small}, \text{Base}, \text{Large}\}$.

\begin{table}[H]
\centering
\footnotesize
\caption{\textbf{FLOPs for MAE models during pre-training on the five considered datasets}. We report the forward GFLOPs (for a single batch element) for both joint-token and token-based multispectral fusion across three model sizes. Multipliers (x) indicate the GFLOPs increase with token-based multispectral fusion compared to joint-token multispectral fusion.}
\label{tab:macs_flops_pretraining}
\begin{tabular}{llllllllllllllll}
\toprule
\multirow{2.6}{*}{\makecell[c]{Multispectral \\ fusion}}
& \multicolumn{3}{c}{\textbf{TreeSatAI-TS}}
& \multicolumn{3}{c}{\textbf{PASTIS-HD}}
& \multicolumn{3}{c}{\textbf{FLAIR\#2}}
& \multicolumn{3}{c}{\textbf{FLAIR-HUB}}
& \multicolumn{3}{c}{\textbf{S2-NAIP urban}} \\
\cmidrule(lr){2-4} \cmidrule(lr){5-7} \cmidrule(lr){8-10} \cmidrule(lr){11-13} \cmidrule(lr){14-16}
& \makecell[c]{Small} & \makecell[c]{Base} & \makecell[c]{Large} 
& \makecell[c]{Small} & \makecell[c]{Base} & \makecell[c]{Large} 
& \makecell[c]{Small} & \makecell[c]{Base} & \makecell[c]{Large} 
& \makecell[c]{Small} & \makecell[c]{Base} & \makecell[c]{Large} 
& \makecell[c]{Small} & \makecell[c]{Base} & \makecell[c]{Large}  \\
\midrule
Joint-token
& 11 & 29 & 80
& 45 & 112 & 308
& 49 & 118 & 320
& 54 & 131 & 355
& 36 & 91  & 252 \\
\rowcolor{black!5}
Token-based 
& 27 & 67 & 188
& 153 & 347 & 891
& 114 & 268 & 705
& 125 & 294 & 778
& 100 & 236 & 635 \\
\midrule
\multirow{2}{*}{\makecell[l]{$\times$ FLOPS \\ w/ token-based}}
& \multirow{2}{*}{$\times$2.4} & \multirow{2}{*}{$\times$2.4} & \multirow{2}{*}{$\times$2.4}
& \multirow{2}{*}{$\times$3.4} & \multirow{2}{*}{$\times$3.1} & \multirow{2}{*}{$\times$2.9}
& \multirow{2}{*}{$\times$2.4} & \multirow{2}{*}{$\times$2.3} & \multirow{2}{*}{$\times$2.2}
& \multirow{2}{*}{$\times$2.3} & \multirow{2}{*}{$\times$2.2} & \multirow{2}{*}{$\times$2.2}
& \multirow{2}{*}{$\times$2.7} & \multirow{2}{*}{$\times$2.6} & \multirow{2}{*}{$\times$2.5} \\\\
\bottomrule
\end{tabular}
\end{table}

\subsection{Probing/Fine-tuning FLOPs}
\label{sm:macs_flops_probing_finetuning}

We reuse the same notation as in \cref{sm:macs_flops_pretraining}. As before, the sequence length is given by $L_m = \left(I_m / P_m\right)^2 D_m$ for \emph{joint-token} multispectral fusion, and by $L_m = \left(I_m / P_m \right)^2 D_m |\mathcal{G}_m|$ for \emph{token-based} multispectral fusion.

For a modality $m$ that is not grouped with others via early multimodal fusion, the MACs in the \emph{encoder} are:
\begin{align*}
\text{MACs}^\text{enc} & = \left( 12 L_m C_e^2 + 2 L_m^2 C_e \right) N_e.
\end{align*}

For the grouped Sentinel-1 ascending and descending modalities, $\text{MACs}^\text{enc}$ is computed by replacing $L_m$ with the \textit{sum of sequence lengths} of the two modalities.

As before, the \emph{patchify} operations contribute:
\begin{align*}
\text{MACs}^\text{patchify} & = \sum_m I_m^2  D_m C_m C_e.
\end{align*}

The attentive pooling operations in the classification and segmentation heads contribute:
\begin{align*}
\text{MACs}^\text{attn-pool} = 2 \sum_m L_m C^2_e.
\end{align*}

The final \emph{dense} projections in the classification and segmentation heads contribute:
\begin{align*}
\text{MACs}^\text{proj-cls} & = C_e C_\text{cls}, \\
\text{MACs}^\text{proj-seg} & = L_\text{ref} C_e C_\text{seg},
\end{align*}
where $L_\text{ref}$ is the sequence length of the spatial \emph{token grid of reference} (see \cref{sec:downstream_tasks}), while $C_\text{cls}$ and $C_\text{seg}$ are the number of semantic classes for the classification and segmentation tasks, respectively, including classes ignored during loss and metric computations (see \cref{sm:datasets}).

Putting everything together, we compute  in \cref{tab:macs_flops_probing_finetuning} the forward FLOPs ($\text{FLOPs} = 2 \times \text{MACs}$) for MAE/ViT models during probing/fine-tuning. FLOPs are computed for both \emph{joint-token} and \emph{token-based} multispectral fusion across three model sizes and the four evaluated datasets. 

The computation is based on the default configurations in \cref{tab:treesat_hyperparams}, \cref{tab:pastis_hyperparams}, and \cref{tab:flair_hyperparams} and the values $N_e=\{12,12,24\}$ and $C_e=\{384,768,1024\}$ for model sizes $\{\text{Small}, \text{Base}, \text{Large}\}$.

\begin{table}[h]
\centering
\footnotesize
\caption{\textbf{FLOPs for MAE/ViT models during probing/fine-tuning on the four evaluated datasets}. We report the forward GFLOPs (for a single batch element) for both joint-token and token-based multispectral fusion across three model sizes. Multipliers (x) indicate the GFLOPs increase with token-based multispectral fusion compared to joint-token multispectral fusion.}
\label{tab:macs_flops_probing_finetuning}
\begin{tabular}{lllllllllllll}
\toprule
\multirow{2.6}{*}{\makecell[c]{Multispectral \\ fusion}}
& \multicolumn{3}{c}{\textbf{TreeSatAI-TS}}
& \multicolumn{3}{c}{\textbf{PASTIS-HD}}
& \multicolumn{3}{c}{\textbf{FLAIR\#2}}
& \multicolumn{3}{c}{\textbf{FLAIR-HUB}} \\
\cmidrule(lr){2-4} \cmidrule(lr){5-7} \cmidrule(lr){8-10} \cmidrule(lr){11-13}
& \makecell[c]{Small} & \makecell[c]{Base} & \makecell[c]{Large} 
& \makecell[c]{Small} & \makecell[c]{Base} & \makecell[c]{Large} 
& \makecell[c]{Small} & \makecell[c]{Base} & \makecell[c]{Large} 
& \makecell[c]{Small} & \makecell[c]{Base} & \makecell[c]{Large}  \\
\midrule
Joint-token
& 21 & 79 & 276
& 95 & 331 & 1125
& 97 & 339 & 1150
& 106 & 375 & 1276 \\
\rowcolor{black!5}
Token-based 
& 52 & 192 & 665
& 374 & 1110 & 3584
& 257 & 816 & 2695
& 277 & 891 & 2954 \\
\midrule
\multirow{2}{*}{\makecell[l]{$\times$ FLOPS \\ w/ token-based}}
& \multirow{2}{*}{$\times$2.5} & \multirow{2}{*}{$\times$2.4} & \multirow{2}{*}{$\times$2.4}
& \multirow{2}{*}{$\times$3.9} & \multirow{2}{*}{$\times$3.4} & \multirow{2}{*}{$\times$3.2}
& \multirow{2}{*}{$\times$2.6} & \multirow{2}{*}{$\times$2.4} & \multirow{2}{*}{$\times$2.3}
& \multirow{2}{*}{$\times$2.6} & \multirow{2}{*}{$\times$2.4} & \multirow{2}{*}{$\times$2.3} \\\\
\bottomrule
\end{tabular}
\end{table}